\documentclass[preprint,12pt]{elsarticle}

\usepackage{amssymb}

\usepackage{bm}

\usepackage{subfig}

\usepackage{amsmath}
\usepackage{graphicx}
\usepackage{my_symbols}
\usepackage{enumitem}
\usepackage{stmaryrd}
\usepackage{tabularx}
\usepackage{verbatim}
\newcommand{\pink}{\color{magenta}}		
\renewcommand{\pink}{}		
\newcommand{\red}{}		
\newcommand{\alg}[1]{\texttt{#1}}

\journal{Pattern Recognition}

\begin{document}

\begin{frontmatter}



\title{Efficient Monte Carlo Methods for Multi-Dimensional Learning with Classifier Chains}


\author[UC3M]{Jesse Read\footnote{Correspoding author, jesse@tsc.uc3m.es.} }
\author[UC3M]{Luca Martino}
\author[UPM]{David Luengo}

\address[UC3M]{Dept.\ of Signal Theory and Communications. Universidad Carlos III de Madrid. Madrid 28911, Spain ({ jesse,luca@tsc.uc3m.es}). }
\address[UPM]{Dept.\ of Circuits and Systems Engineering. Universidad Polit\'{e}cnica de Madrid. Madrid 28031, Spain ({ david.luengo@upm.es}).}

\begin{abstract}
Multi-dimensional classification (MDC) is the supervised learning problem where an instance is associated with multiple classes, rather than with a single class, as in traditional
 classification problems.
%
%
Since these classes are often strongly correlated, modeling the dependencies between them allows MDC methods to improve their performance -- at the expense of an increased computational cost.
In this paper we focus on the classifier chains (CC) approach for modeling dependencies, one of the most popular and highest-performing methods for multi-label classification (MLC), a particular case of MDC which involves only binary classes (i.e., labels).
The original CC algorithm makes a greedy approximation, and is fast but tends to propagate errors along the chain.
%
%
Here we present novel Monte Carlo schemes, both for finding a good chain sequence and performing efficient inference.
%
%
Our algorithms remain tractable for high-dimensional data sets and obtain the best predictive performance across several real data sets.
\end{abstract}


\end{frontmatter}


\section{Introduction}
\label{sec:Intro}

Multi-dimensional classification (MDC) is the supervised learning problem where an instance may be associated with multiple classes, rather than with a single class as in traditional binary or multi-class {\pink{single}}-dimensional classification (SDC) problems.
So-called MDC (e.g., in \cite{Bielza2011}) is also known in the literature as multi-target, {\pink{multi-output} \cite{RFPCT}, or multi-objective \cite{MODTs}} classification\footnote{Multi-output, multi-target, multi-variate etc.\  can also refer to the regression case, where the outputs are continuous}, and is related to multi-task clustering and multi-task learning.
The recently popularised task of multi-label classification (see \cite{Overview,Carvalho2009,Thesis,MMD} for overviews) can be viewed as a particular case of the multi-dimensional problem that only involves binary classes, i.e., \key{labels} that can be turned on $(1)$ or off $(0)$ for any data instance.
The MDC learning context is receiving increased attention in the literature, since it arises naturally in a wide variety of domains, such as image classification \cite{Scene,Qi2009}, information retrieval and text categorization \cite{BPMLL}, automated detection of emotions in music \cite{Music} or bioinformatics \cite{BPMLL,GeneFunction}.

The main challenge in this area is modeling label dependencies while being able to deal with the scale of real-world problems.
A basic approach to MDC is the independent classifiers (\texttt{IC}) method, (commonly known as \textit{binary relevance} in multi-label circles), which decomposes the MDC problem into a set of SDC problems (one per label) and uses a separate classifier for each label variable.\footnote{Throughout this work we use the term \key{label} to refer generally to a \key{class variable} that takes a number of discrete values (i.e., classes); not necessarily binary as in the multi-label case}
In this way, MDC is turned into a series of standard SDC problems that can be solved with any off-the-shelf binary classifier (e.g., a logistic regressor or a support vector machine\footnote{Support vector machines (SVMs) are naturally binary, but can be easily adapted to a multi-class scenario by using a pairwise voting scheme, as in \cite{SMOM}}).
Unfortunately, although \texttt{IC} has a low computational cost, it obtains unsatisfactory performance on many data sets {\pink{and performance measures,}} because it does not take into account the dependencies between labels \cite{Thesis,RAKEL,ECC2,PCC,GuoGu,UPM}.

In order to model dependencies explicitly, several alternative schemes have been proposed, such as the so-called \emph{label powerset} (\texttt{LP}) method \cite{Overview}.
\texttt{LP} considers each potential combination of labels in the MDC problem as a single label.
In this way, the multi-dimensional problem is turned into a traditional multi-class SDC problem that can be solved using standard methods.
Unfortunately, given the huge number of class values produced by this transformation (especially for non-binary labels), this method is usually unfeasible for practical application, and suffers from issues like overfitting.
This was recognised by \cite{RAKEL,EPS}, which provide approximations to the \texttt{LP} scheme that reduce these problems, although such methods have been superseded in recent years (as shown in \cite{ExtML}).

A more recent idea is using classifier chains (\texttt{CC}), which improves the performance of \texttt{IC} and \texttt{LP} {\pink{on some measures (e.g., the subset 0/1 loss)}} by constructing a sequence of classifiers that make use of previous outputs of the chain {\pink(see \cite{OnLabelDependenceJournal} for a detailed discussion on MLC methods and loss functions).}
The original \texttt{CC} method \cite{ECC2} performs a greedy approximation, and is fast (similar to \texttt{IC} in terms of complexity) but is susceptible to error propagation along the chain of classifiers.
Nevertheless, a very recent extensive experimental comparison reaffirmed that \texttt{CC} is among the highest-performing methods for MLC, and recommended it as a benchmark algorithm \cite{ExtML}.

A \texttt{CC}-based Bayes-optimal method, probabilistic classifier chains (\texttt{PCC}), was recently proposed \cite{PCC}.
However, although it improves the performance of \texttt{CC}, its computational cost is too large for most real-world applications.
Some approaches have been proposed to reduce the computational cost of PCC at test time \cite{UPM,BeamSearch,ChainingMLC}, but the problem is still open.
Furthermore, the performance of all \texttt{CC}-based algorithms depends on the label order established at training time, an issue that so far has only been considered by \cite{BeamSearch} using a heuristic search algorithm called beam search.

In this paper we introduce novel methods that attain the performance of \texttt{PCC}, but remain tractable for high-dimensional data sets both at training and test times.
Our approaches are based on a double Monte Carlo optimization technique that, aside from tractable inference, also explicitly searches the space of possible chain-sequences during the training stage.
Another advantage of the proposed algorithms is that predictive performance can be traded off for scalability depending on the application.
Furthermore, we demonstrate our methods with support vector machine (SVM) {\red{as base}} classifiers ({\red PCC methods have only been used under a logistic regression scheme so far}).
Finally, unlike the bulk of related literature, we involve the general multi-dimensional scenario (as in \cite{UPM, RFPCT}) and provide a theoretical and empirical analysis of payoff functions for searching the chain space.

A preliminary version of this work has been published in \cite{MCC}.
With respect to that paper, here we introduce three major improvements: we consider the more challenging scenario of multi-dimensional classification (i.e., multi-class labels); at the training stage, we address the problem of finding the optimum label order instead of accepting the original one or using a random label order; for the test stage, we develop a more sophisticated and efficient population Monte Carlo approach for inference.

The paper is organized as follows.
In the following \Sec{sec:MLC} we review MDC and the important developments leading up to this paper.
In \Sec{sec:training} and \Sec{sec:test} we detail our novel methods for training (including learning the optimum chain sequence) and inference, respectively.
In \Sec{sec:Exp} we elaborate an empirical evaluations of the proposed algorithms and , finally, in \Sec{sec:Concl} we draw some conclusions and mention possible future work.

\section{Multi-Dimensional Classification (MDC)}
\label{sec:MLC}


Let us assume that we have a set of training data composed of $N$ labelled examples, $\D = \{(\x^{(n)},\y^{(n)})\}_{n=1}^N$,  where 
\begin{equation*}
	\x^{(n)} = [x_1^{(n)},\ldots,x_D^{(n)}]^{\top}
		\in {\boldsymbol \X} = \mathcal{X}_1 \times \cdots \times \mathcal{X}_D \subseteq \mathbb{R}^D
\end{equation*}
is the $n$-th feature vector (input), and
\begin{equation*}
	\y^{(n)} = [y_1^{(n)},\ldots,y_L^{(n)}]^{\top}
		\in {\boldsymbol \Y} = \mathcal{Y}_1 \times \cdots \times \mathcal{Y}_L \subset \mathbb{N}_+^L
\end{equation*}
is the $n$-th label vector (output), with
\begin{equation*}
	y^{(n)}_\ell \in \mathcal{Y}_\ell = \{1,\ldots,K_\ell \},
\end{equation*}
and $K_\ell\in \mathbb{N}_+$ being the finite number of classes associated to the $\ell$-th label. 
The goal of MDC is learning a classification function,\footnote{We consider $\h$ as a vector because this fits naturally into the independent classifier and classifier chain context, but this is not universal, and $h : {\boldsymbol \X} \rightarrow {\boldsymbol \Y}$ is possible in other contexts (such as \texttt{LP})}
\begin{equation*}
	\h = [h_1,\ldots,h_L]^{\top} : {\boldsymbol \X} \rightarrow {\boldsymbol \Y}.
\end{equation*}

Let us assume that the unknown {\it true} posterior {\red{probability density function (PDF)}} of the data is $p(\y|\x)$.
From a Bayesian point of view, the optimal label assignment for a given test instance, $\xtest$, is provided by the maximum a posteriori (MAP) label estimate,
\begin{equation}
	\hat{\y}_{\map} = \h_{\map} (\xtest) = \argmax_{\y {\red{\in {\boldsymbol \Y}}}} p(\y|\xtest),
\label{eq:optMapMlc}
\end{equation}
where the search must be performed over all possible test labels, $\y \in {\boldsymbol \Y}$.
{\pink{The MAP label estimate is the one most commonly used in the literature, although other approaches are possible, as shown in \cite{PCC}.
Indeed, \cite{PCC} shows that \Eq{eq:optMapMlc} minimizes the \key{exact match} or \key{subset 0/1} loss, whereas the \key{Hamming} loss is minimized by finding individual classifiers that maximize the conditional probability for each label.}}
Unfortunately, the problem is further complicated by the fact that the true {\red{density}}, $p(\y|\x)$, is usually unknown, and the classifier has to work with an approximation, $\hat{p}(\y|\x)$, constructed from the training data.
Hence, the (possibly sub-optimal) label prediction is finally given by
\begin{equation}
	\yest = \h(\xtest) = \argmax_{\y {\red{\in {\boldsymbol \Y}}}}\ \hat{p}(\y|\xtest).
\label{eq:suboptMapMlc}
\end{equation}
Table \ref{TableNot} summarizes the main notation used throughout this work.

\begin{table}[!hbt]
\small
\begin{center}
\caption{Summary of the main notation used in this work.}
\label{TableNot}
\begin{tabularx}{\textwidth}{lX}
\toprule
{\bf Notation} & {\bf Description}  \\
	\midrule
	$\x = [x_1,\ \ldots,\ x_D]^{\top} \in {\boldsymbol \X}\subseteq \mathbb{R}^D$ & $D$-dimensional feature/input vector, with $x_d
		\in \X_d \subseteq \mathbb{R}$, $d=1,\ldots,D$. \\
	$\y = [y_1,\ldots,y_L]^{\top} \in {\boldsymbol \Y}\subset \mathbb{N}_+^L$ & $L$-dimensional label/output vector, with $y_\ell
		\in \mathcal{Y}_\l = \{1,\ldots,K_\l \}$ ($K_\l\geq 2$), $\l=1,\ldots,L$. \\
	$\mat{X} = [\x^{(1)},\ \ldots,\ \x^{(N)}] \in {\boldsymbol \X}^N$ & $D \times N$ input matrix with all the features. \\
	$\mat{Y} = [\y^{(1)},\ \ldots,\ \y^{(N)}] \in {\boldsymbol \Y}^N$ & $L \times N$ output matrix with all the labels. \\
	$\D = \{(\x^{(n)},\y^{(n)})\}_{n=1}^N = \{\mat{X},\mat{Y}\}$ & Training data set, $n=1,\ldots,N$.\\
  $p(\y|\x)$ & Unknown true {\red{PDF}} of the data. \\
  $\hat{p}(\y|\x)$ & Empirical {\red{PDF}} built by the classifier. \\
	$\xtest {\red{= [x_1^*,\ \ldots,\ x_D^*]^{\top} \in {\boldsymbol \X}}}$ & Test feature vector.\\
	$\h = [h_1,\ \ldots,\ h_L]^{\top}: {\boldsymbol \X} \rightarrow {\boldsymbol \Y}$ & Classification function built from $\D$.\\	
	$\hat{\y} = \h(\x) = [\hat{y}_1,\ \ldots,\ \hat{y}_L]^{\top}$ & Generic classifier's output. \\
	$\hat{\mat{Y}} = \mat{H}(\mat{X}): {\boldsymbol \X}^N \rightarrow {\boldsymbol \Y}^N$ & Classification matrix $\mat{H}$ applied to 
		$\mat{X}$. \\
	$\s = [s_1,\ \ldots,\ s_L]^{\top}\in \S^L$  & Label order, with $s_\l \in \S = \{1,\ldots,L\}$.\\
	$\y_{\s} = [y_{s_1},\ \ldots,\ y_{s_L}]^{\top} \in {\boldsymbol \Y}$ & $L$-dimensional permuted label vector.\\
	$\h_{\s} = [h_{s_1},\ \ldots,\ h_{s_L}]^{\top}:  {\boldsymbol \X} \rightarrow {\boldsymbol \Y} $ & Permuted classification function.\\
\bottomrule
\end{tabularx}
\end{center}
\end{table} 

\subsection{\red{Multi-Dimensional Classification vs.\ Multi-Label Classification}}
\label{sec:MDvML}

Although binary-only multi-label problems can be considered as a subset of multi-dimensional problems, the reverse is not true, and there are some important quantitative and qualitative differences.
Quantitatively, there is a higher dimensionality (for the same value of $L$); MLC deals with $2^L$ possible values, whereas MDC deals with $\prod_{\l=1}^L K_{\l}$.
Note that this affects the inference space, but \emph{not} the sequence space (i.e., the possible orderings of variables).
Qualitatively, in MDC the distribution of ``labellings'' is different, even with binary class variables.
In typical MLC problems, the binary classes indicate \emph{relevance} (e.g., the label \textsf{beach} is relevant (or not) to a particular image).
Hence, in practice only slightly more than $1/L$ labels are typically relevant to each example on average \cite{Thesis} (see also \Tab{table:datasets}), i.e., $\sum_{\ell=1}^L P(y_\ell) \ll L$ {\red{where $P(y_\ell)$ is the probability of $y_\ell$ being relevant.}}
This means that a relatively small part of the $\Y$-space is used.
In MDC, classes (including binary classes) are used differently -- e.g., a class \textsf{gender}  ($\in \{1,2\} \equiv$ \textsf{M}/\textsf{F}) -- with a less-skewed distribution of classes; prior-knowledge of the problem aside, we expect $P(Y_\ell=y_\ell) \approx \frac{1}{K_\ell}$.
In summary, in MDC the \emph{practical} $\Y$-space is much greater than in MLC, making probabilistic inference more challenging.

\vspace*{6pt}

\subsection{Independent Classifiers (\texttt{IC})} 

The method of using \emph{independent classifiers} (\texttt{IC}) on each label is commonly mentioned in the MLC and MDC literature \cite{Overview,MMD,ECC2,UPM}.
%
%
For each $\l=1,\ldots,L$ a [standard, off-the-shelf binary] classifier $h_{\l}$ is employed to map new data instances to the relevance of the $\l$-th label, i.e.,
\begin{equation*}
	\yest = \h(\x^*) = [h_1(\x^*),\ldots,h_L(\x^*)]^{\top},
\end{equation*}
where, probabilistically speaking, we can define each $h_{\l}$ as
\begin{equation}
	 \hat{y}_{\l} = h_{\l}(\x^*) = \argmax_{y_{\l} {\red{\in \mathcal{Y}_\l}}}\ \hat{p}(y_{\l}|\x^*).
\label{eq:ic}
\end{equation}
As we remarked in \Sec{sec:Intro}, this method is easy to build using off-the-shelf classifiers, but it does not explicitly model label dependencies, and its performance suffers as a result.\footnote{\pink{An exception to this rule is the minimization of the Hamming loss, which can be attained by considering each of the individual labels separately. Thus, modeling label dependencies does not provide an advantage in this case, as already discussed in \cite{PCC,OnLabelDependenceJournal}}}
In fact, it assumes complete independence, i.e., it approximates the {\red{density}} of the data as
\begin{equation}
	\hat{p}(\y|\x) = \prod_{\l=1}^L \hat{p}(y_{\l}|\x).
\end{equation}
We always expect label dependencies in a multi-label problem (otherwise we are simply dealing with a collection of unrelated problems); some labels occur more likely together, or mutually exclusively.
Thus, it is important to model these dependencies, because doing so can greatly influence the outcome of the predictions.

\subsection{Classifier Chains (\texttt{CC})} 

\begin{figure*}[!tb]
\begin{center}
\includegraphics[width=\textwidth]{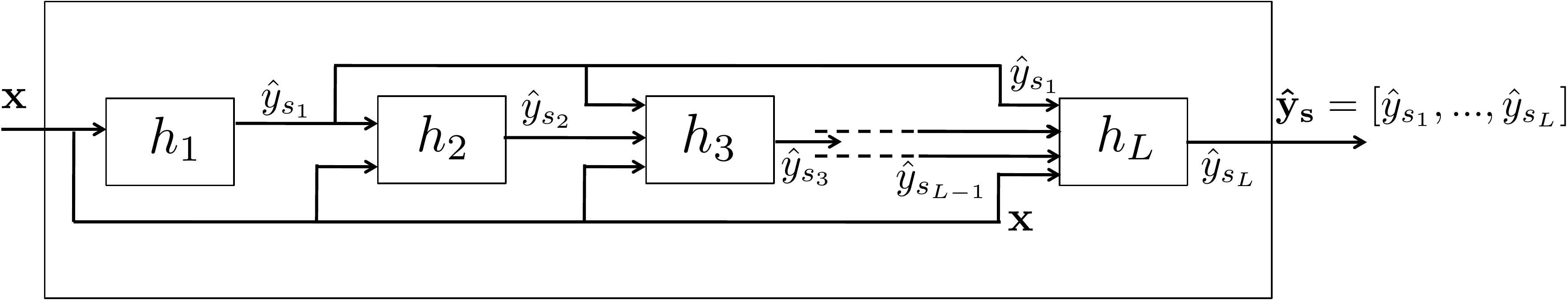}
\caption{General scheme of the Classifier Chains (CC) approach.}
\label{Fig2}
\end{center}
\end{figure*}

\begin{figure}[!tb]
\begin{center}
 \includegraphics[width=0.5\textwidth]{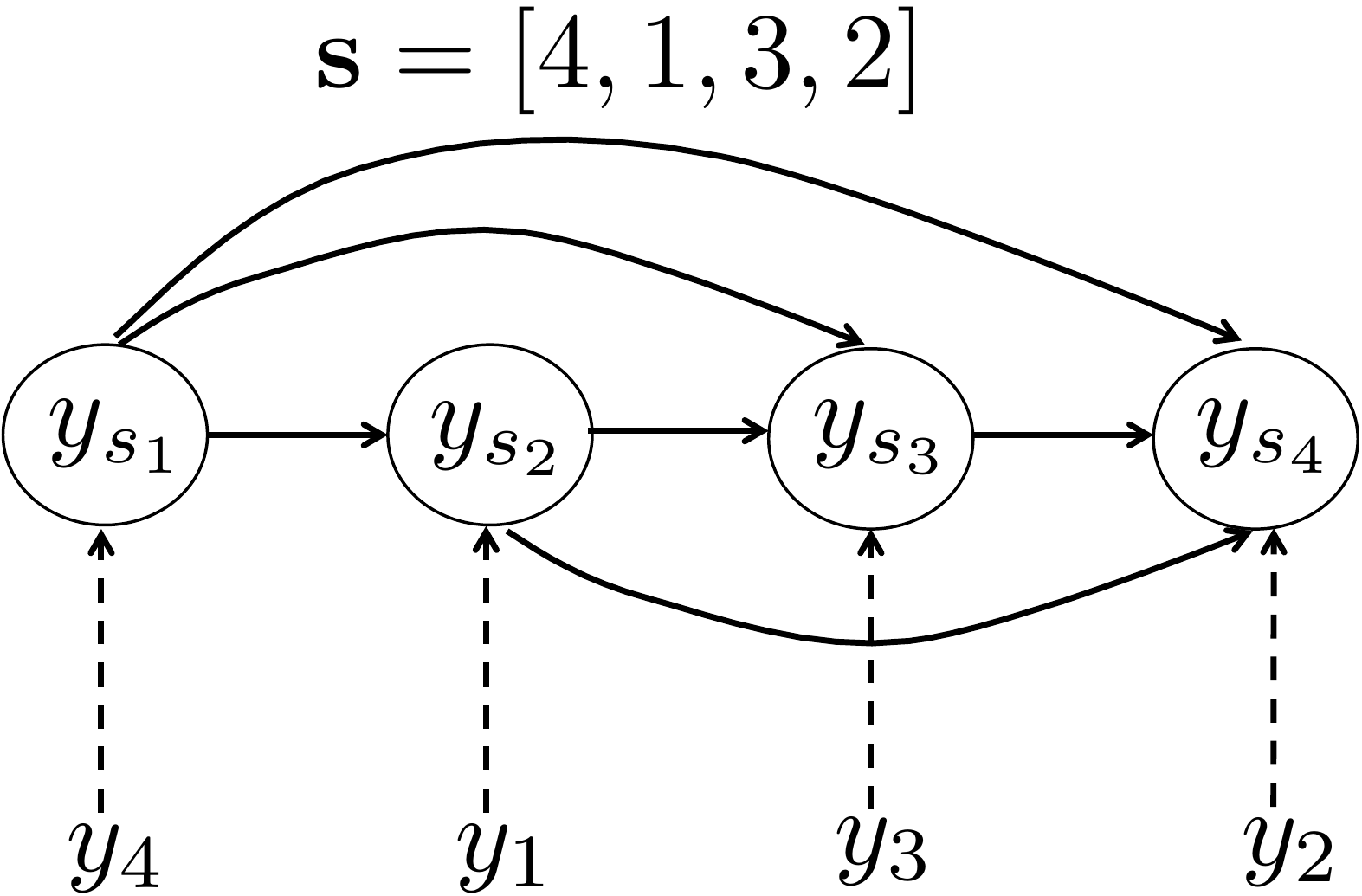}
\caption{Example of the permuted label vector in a classifier chain with $L=4$. In this example we have $\s=[4,1,3,2]^{\top}$, so that $\y_\s = [y_4,y_1,y_3,y_2]^{\top}$.}
\label{Fig1}
\end{center}
\end{figure}

The \key{classifier chains} (\texttt{CC}) approach \cite{ECC2} is based on modeling the correlation among labels using the chain rule of probability (see Figure \ref{Fig2}).
Given a test instance, $\xtest$, the true label probability may be expressed exactly as
\begin{equation}
	p(\y|\xtest) = p(y_1|\xtest) \prod_{\l=2}^{L}{p(y_{\l}|\xtest,y_{1},\ldots,y_{\l-1})},
\label{eq:trueChain}
\end{equation}
{\red{Theoretically}}, label order is irrelevant {\red{in \Eq{eq:trueChain}}}, as all the label orderings result in the same {\red{PDF}}.
However, since in practice we are modelling an approximation of $p$ (i.e., $\hat{p}$), label order can be very important for attaining a good classification performance, as recognized in \cite{PCC,OnLabelDependenceJournal}.
Given some label order, $\s = [s_1,\ldots,s_L]^{\top}$ (a permutation of $\{1,\ldots,L\}$), \texttt{CC} approximates the true data {\red{density}} as 
\begin{equation}
	\hat{p}(\y_{\s}|\xtest) = \hat{p}(\y|\xtest,\s)
		= \hat{p}(y_{s_1}|\xtest) \prod_{\l=2}^{L}{\hat{p}(y_{s_{\l}}|\xtest,y_{s_1},\ldots,y_{s_{\l-1}})}, 
\label{eq:chain}
\end{equation}
where $\y_{\s} = [y_{s_1},\ \ldots,\ y_{s_L}]^{\top}$ is the permuted label vector (see Figure \ref{Fig1}).

\begin{figure*}[!tb]
\begin{center}
	\includegraphics[width=0.75\textwidth]{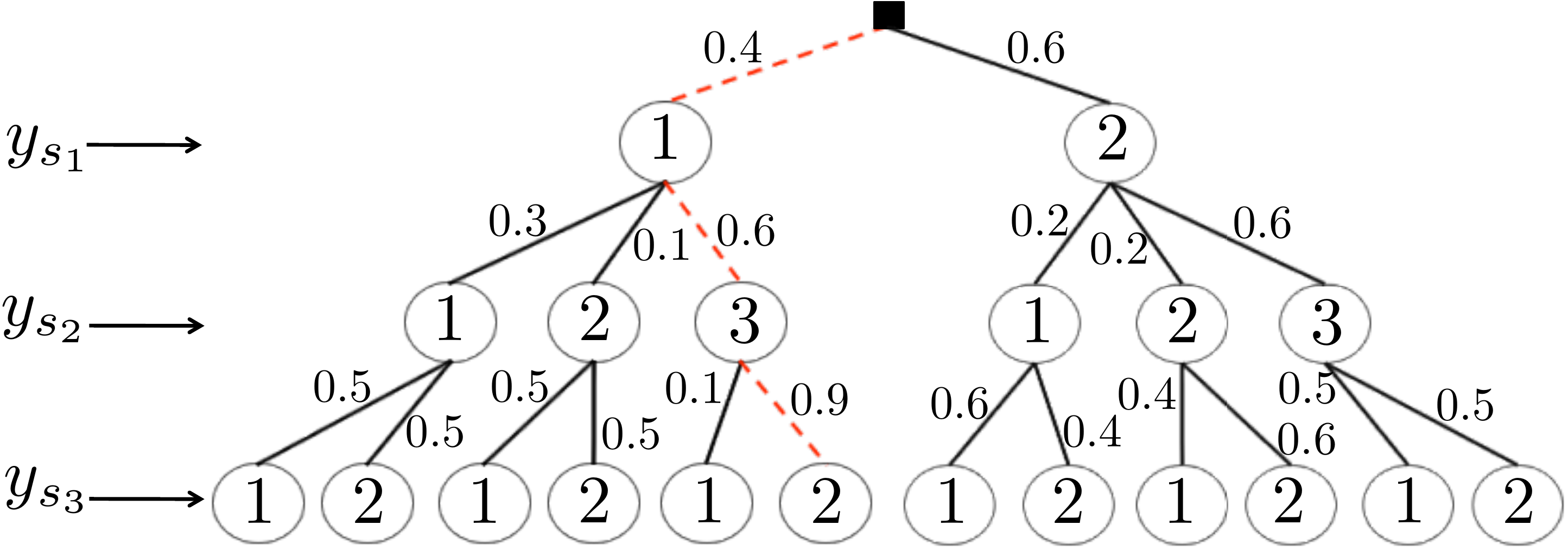}
\caption{Example of the $\prod_{\l=1}^{L} K_\l = K_1 \times K_2 \times K_3 = 2 \times 3 \times 2 = 12$ possible paths along the tree of class labels $y_{s_\l}$ ($\l=1,\ldots,L=3$). The best path, $\y_\s = [1,3,2]^{\top}$, with probability $0.2160$, is shown with dashed lines.}
\label{Fig3}
\end{center}
\end{figure*}

First of all, \texttt{CC} considers an arbitrary label order, $\s$, and learns all the conditional probabilities in \eqref{eq:chain} from the labelled data during the training stage, thus effectively constructing a chain of classifiers like the one shown in Figure \ref{Fig2}.
Then, during the test stage, given a new (test) instance, $\xtest$, \texttt{CC} predicts $\hat{y}_{s_1} = h_{s_1}(\xtest)$ using only the feature vector, whereas for the $\l$-th permuted label ($\l=2,\ldots,L$) it also makes use of all the previous predictions ($\hat{y}_{s_1},\ldots,\hat{y}_{s_{\l-1}}$), predicting each $\hat{y}_{s_{\l}}$ as
\begin{equation}
\label{eq:cc}
	\hat{y}_{s_{\l}} = h_{s_{\l}}(\xtest|\hat{y}_{s_1},\ldots,\hat{y}_{s_{\l-1}})
		= \argmax_{y_{s_{\l}} {\red{\in \mathcal{Y}_{s_\l}}}}\ \hat{p}(y_{s_{\l}}|\xtest,\hat{y}_{s_1},\ldots,\hat{y}_{s_{\l-1}}).
\end{equation}
Note that, given a data instance $\xtest$ and a label order $\s$, each possible realization of the vector $\y_{\s}$ can be seen as a path along a tree of depth $L$, and $\hat{p}(\y_{\s}|\xtest)$ is the \key{payoff} or utility corresponding to this path.
\texttt{CC} follows a single path of labels $\y_{\s}$ greedily down the chain of $L$ binary classifiers, as shown in Figure \ref{Fig3} through a simple example.
In carrying out classification down a chain in this way, \texttt{CC} models label dependencies and, as a result, usually performs much better than \texttt{IC}, while being similar in memory and time requirements in practice.
However, due to its greedy approach (i.e., only one path is explored) and depending on the choice of $\s$, its performance can be very sensitive to errors, especially in the initial links of the chain \cite{PCC}.

\subsection{Probabilistic Classifier Chains (\texttt{PCC}) and extensions} 
\label{sec:pcc}

\key{Probabilistic classifier chains} (\texttt{PCC}) was introduced in \cite{PCC}.
In the training phase, \texttt{PCC} is identical to \texttt{CC}; considering a particular order of labels $\s$ (either chosen randomly, or as per default in the dataset).
However, during the test stage \texttt{PCC} provides Bayes-optimal inference by exploring all the $\prod_{\l=1}^{L} K_{\l}=2^{L}$ possible paths (note that \cite{PCC} only considers the MLC case, where $K_{\l}=2$ for $\l=1,\ldots,L$).
Hence, for a given test instance, $\xtest$, \texttt{PCC} provides the optimum $\y_\s$ {\pink{that minimizes the subset 0/1 loss}} by maximizing the probability of the complete label vector, rather than the individual labels (as in \Eq{eq:cc}), i.e.,
\begin{equation}
	\yest_{\s} = \vec{h}_{\s}(\xtest) = \argmax_{\y_{\s} {\red{\in {\boldsymbol \Y}}}}\ \hat{p}(\y_{\s}|\xtest),
\label{eq:pcc}
\end{equation}
where $\hat{p}(\y_{\s}|\xtest)$ is given by \eqref{eq:chain}.\footnote{\pink{Interestingly, it has been shown in \cite{PCC,OnLabelDependenceJournal} that the optimum set of labels that minimize the Hamming loss is given by \eqref{eq:ic}, i.e., the \texttt{IC} approach is optimal for the Hamming loss and no gain is to be expected from any other method that models correlation among labels.}}
In \cite{PCC} an overall improvement of \texttt{PCC} over \texttt{CC} is reported, but at the expense of a high computational complexity: it is intractable for more than about $10$ labels ($\equiv 2^{10}$ paths), which represents the majority of practical problems in the multi-label domain.
Moreover, since all the conditional densities in \eqref{eq:chain} are estimated from the training data, the results can also depend on the chosen label order $\s$, as in \texttt{CC}.

{\pink{An approximate \texttt{PCC}-based inference method with a reduced computational cost has been proposed in \cite{ChainingMLC}.
This approach, named $\epsilon$-approximate inference, is based on performing a depth-first search in the probabilistic tree with a cutting-off list.
It is characterized by quite strong theoretical guarantees regarding the worst-case regret for the subset 0/1 loss and shows a good performance in the experiments, but does not tackle the chain ordering problem.
An alternative approach, `beam search' for \texttt{PCC}, has been proposed in \cite{BeamSearch}.
Beam search is a heuristic search algorithm that speeds up inference considerably and also allows experimentation with chain orderings.
Furthermore, the authors of \cite{BeamSearch} mention the (promising) possibility of using Monte Carlo methods in future works.
A simple Monte Carlo-based \texttt{PCC} approach has been considered in \cite{ChainingMLC,FmeasureMLC} for maximization of the Hamming and the F-measure loss functions respectively during the test (i.e., inference) stage.
We have independently developed a Monte Carlo-based approach in \cite{MCC}, which considers not only the test stage but also the training (i.e., chain order optimization) stage.
In this paper we elaborate on this work, providing more sophisticated Monte Carlo algorithms that speed up both the training and test stages.}}

\subsection{Bayesian Network Classifiers}
\label{sec:cdn}

Conditional dependency networks (\texttt{CDN}) \cite{GuoGu} are used as a way of avoiding choosing a specific label order $\s$.
Whereas both \texttt{CC} and \texttt{PCC} are dependent on the order of labels appearing in the chain, \texttt{CDN} is a fully connected network comprised of $L$ label-nodes $\hat{p}(y_{\l}|\x,y_1,\ldots,y_{\l-1},y_{\l+1},\ldots,y_L)$ for $\l=1,\ldots,L$.
Gibbs sampling is used for inference over $T$ steps, and the marginal probabilities collected over the final $T_c$ steps.
However, due to having $L(L-1)/2$ links, inference may not scale to large $L$.

Bayesian Classifier Chains \cite{UPM} finds a more tractable (non fully-connected) network based on a maximum spanning tree of label dependencies; although they again use the faster classifier chain-type inference, i.e., by treating the resulting graph as a directed one (by electing one of the nodes to be a root, and thus turning the graph into a tree).
%
%
This method is similar to \texttt{CC} in the sense that classification depends on the order of nodes, but, unlike \texttt{CC}, it does not model all dependencies (e.g., the dependence between leaf variables is not necessarily modelled).

\subsection{Inference in MDC: our approach}

As explained in the previous sections, the optimal solution to the classifier chain problem is twofold: 
\begin{enumerate}
	\item Find the best label order $\s$, exploring all the $L!$ possible label orders.
	\item Find the best label vector $\y_\s$ within a space composed of $\prod_{\l=1}^{L} K_{\l}$ possible label vectors.
\end{enumerate}
Unfortunately, this task is unfeasible except for very small values of $L$ and $K_{\l}$ ($\l=1,\ldots,L$).
Indeed, the total space has a cardinality $(\prod_{\l=1}^L K_{\l}) \times L!$ (i.e., exponential times factorial).
For this reason, in the following we design efficient Monte Carlo techniques to provide good solutions to both problems: finding a good label order $\s$ at the training stage (see \Sec{sec:training}), and then a good label vector $\y$ at the test (i.e., inference) stage (see \Sec{sec:test}).

\section{Training stage: Finding the best classifier chain}
\label{sec:training}

In the training step, we want to learn each of the individual classifiers, $h_{s_\ell}$ for $\ell=1,\ldots,L$, and, at the same time, we also wish to find the best chain order, $\s = [s_1,\ldots,s_L]^{\top}$, out of the $L!$ possibilities. We use a Monte Carlo approach to search this space efficiently.

\subsection{Learning the label order}
\label{sec:MonteCarlo}

A first, simple, exploration of the label-sequence space is summarized in Algorithm \ref{alg:full}.
This algorithm can start either with a randomly chosen label order or with the default label order in the dataset, $\s_0$.
In each iteration a new candidate sequence $\s'$ is generated randomly according to a chosen proposal density $\pi(\s|\s_{t-1})$ (see \Sec{sec:proposal} for further details).
Then, a suitable payoff function $J(\s')$ is evaluated (see Section \ref{sec:trainingProblem} for a discussion on possible payoff functions).
The new candidate label order, $\s'$, is accepted if the value of the payoff function is increased w.r.t.\ the current one, $\s_{t-1}$ (i.e., if $J(\s') > J(\s_{t-1})$, then $\s_t = \s'$).
Otherwise, it is rejected and we set $\s_t = \s_{t-1}$.
After a fixed number of iterations $T_s$, the stored label order $\s_{T_s}$ is returned as the output of the algorithm, i.e., the estimation of the best chain order provided is $\spred = \s_{T_s}$.\footnote{In order to avoid overfitting, this is typically performed using internal train/test split or cross validation, i.e., using part of the training set for building the model and the rest for calculating its payoff. See \Sec{sec:Exp} for further details.}

\begin{algorithm}[!tb]
\caption{\label{alg:full} Finding a good label order $\spred$}
Input:
\begin{itemize}[noitemsep,nolistsep]
	\item $\D = \{(\x^{(n)},\y^{(n)})\}_{n=1}^{D}$: training data.
	\item $\pi(\s|\s_{t-1})$: proposal density.
	\item $\s_0,T_s$: initial label order and number of iterations.
\end{itemize}
Algorithm:
\begin{enumerate}[noitemsep,nolistsep]
	\item For $t=1,\ldots,T_s$:
	\begin{enumerate}[noitemsep,nolistsep]
		\item Draw $\s' \sim \pi(\s|\s_{t-1})$.
		\item \texttt{if} $J(\s') > J(\s_{t-1})$
		\begin{itemize}[noitemsep,nolistsep]
			\item $\s_t \gets \s'$ accept.
		\end{itemize}
		\item \texttt{else}
		\begin{itemize}[noitemsep,nolistsep]
			\item $\s_t \gets \s_{t-1}$ reject.
		\end{itemize}
	\end{enumerate}
\end{enumerate}
Output:
\begin{itemize}[noitemsep,nolistsep]
	\item $\spred = \s_{T_s}$: estimated label order.
\end{itemize}
\end{algorithm}

As we show in Section \ref{sec:trainingProblem}, the payoff function $J(\s)$ is based on $\hat{p}(\y_{\s}|\x)$, an approximation of the true data {\red{density}}, $p(\y_{\s}|\x)$.
Hence, in order to decrease the dependence on the training step we can consider a population of estimated label orders, $\mat{S} = [\spred^{(1)},\ldots,\spred^{(M)}]$, instead of a single one.
This method is detailed in \Code{alg:full2}. The underlying idea is similar to the previous \Code{alg:full}, but returning the best $M$ label orders (the ones with the highest payoff) after $T_s$ iterations instead of a single label order.

\begin{algorithm}
\caption{\label{alg:full2} Finding a good population of label orders $\mat{S} = [\spred^{(1)},\ldots,\spred^{(M)}]$}
Input:
\begin{itemize}[noitemsep,nolistsep]
  \item $\D = \{(\x^{(n)},\y^{(n)})\}_{n=1}^{D}$: training data.
	\item $\pi(\s|\s_{t-1})$: proposal density.
	\item $\s_0,T_s,M$: initial order, number of iterations and population size.
\end{itemize}
Algorithm:
\begin{enumerate}[noitemsep,nolistsep]
	\item For $t=1,\ldots,T_s$:
	\begin{enumerate}[noitemsep,nolistsep]
		\item Draw $\s' \sim \pi(\s|\s_{t-1})$.
		\item \texttt{if} $J(\s') \geq J(\s_{t-1})$
		\begin{itemize}[noitemsep,nolistsep]
			\item $\s_{t} \gets \s'$ accept.
			\item $w_t \gets  J(\s')$ set.
		\end{itemize}
		\item \texttt{else} 
		\begin{itemize}[noitemsep,nolistsep]
			\item $\s_{t} \gets \s_{t-1}$ accept.
			\item $w_t \gets  J(\s_{t-1})$ set.
		\end{itemize}
	\end{enumerate}
	\item Sort $\s_1,\ldots,\s_{T_s}$ decreasingly w.r.t.\ $w_1,\ldots,w_{T_s}$, taking the top $M$.
\end{enumerate}
Output:
\begin{itemize}[noitemsep,nolistsep]
	\item $\mat{S} = [\spred^{(1)},\ldots,\spred^{(M)}]$: population of best $M$ estimated label orders.
	\item $w^{(1)},\ldots,w^{(M)}$: corresponding weights.
\end{itemize}
\end{algorithm}

Once we have described the two proposed Monte Carlo approaches for the training step, the following two sections are devoted to the critical issues for both of the algorithms: the choice of the proposal (\Sec{sec:proposal}) and of the payoff function (\Sec {sec:trainingProblem}).

\subsection{Choice of the proposal function}
\label{sec:proposal}

In order to explore the sequence space, $\S$, a proposal mechanism is required.
We remark that performing a search in $\S$ requires {\bf (a)} learning a probabilistic model and {\bf (b)} building a new classifier chain for each sequence we want to try.
Hence, this stage is inherently much more expensive than searching the label space and the number of label orders that can be explored is thus very limited.
Therefore, the proposal density must be both simple and effective. Below, we describe two possibilities. 

{\it First proposal scheme:} As a first approach we consider a very simple proposal. Specifically, given a sequence 
$$
	\s_{t-1} = [s_{t-1}(1),\ldots,s_{t-1}(L)]^{\top},
$$ 
the proposal function $\pi(\s_t|\s_{t-1})$ consists of choosing uniformly two positions of the label order  ($1 \le \ell,m \le L$) and swapping the labels corresponding to those positions, so that $\s_t(\ell) = \s_{t-1}(m)$ and $\s_t(m) = \s_{t-1}({\red\ell})$.

{\it Second proposal scheme:} The previous proposal does not make a full use of all the available information.
For instance, due to the chain structure, changing the initial `links' in the chain (e.g., $s_t(1)$ or $s_t(2)$) implies a larger jump in the sequence space than changing the final links (e.g., $s_t(L-1)$ or $s_t(L)$).
Indeed, if the first $L_1$ labels in $\s_t$ remain unchanged w.r.t.\ $\s_{t-1}$, only $L-L_1$ classifiers need to be re-trained, thus saving valuable computation time.
In light of this observation, we propose an improvement of the previous proposal based on freezing the links in the chain progressively from the beginning to the end.\footnote{This idea follows the line of the different tempering strategies found in the literature, such as simulated annealing or simulated tempering \cite{SA}. However, from a Monte Carlo point of view there is an important difference: our tempering is applied to the proposal, whereas the classical tempering is used to change the target.}
This allows the algorithm to explore the whole sequence space uniformly in the $T_p$ initial iterations (i.e., potentially requiring re-training of the whole classifier chain), but focuses gradually on the last labels of the sequence, which require almost no re-training and are very cheap to explore.
In this case, the first label at the $t$-th iteration is drawn from
\begin{equation}
\label{Eqpesos1}
	p_{{\red\ell},t} \propto
	\begin{cases}
		\frac{1}{N}, & t \le T_p;\\
		\left(\frac{1}{N}\right)^{\frac{\beta t}{\red{\ell}}}, & t > T_p;
	\end{cases}
\end{equation}
with the second label drawn from
\begin{equation}
\label{Eqpesos2}
	p_{{\red{m}},t} \propto
	\begin{cases}
		\frac{1}{N-1}, & t \le T_p;\\
		\left(\frac{1}{N-1}\right)^{\frac{\beta t}{\red{m}}}, & t > T_p,
	\end{cases}
\end{equation}
where $\beta > 0$ is a user-defined and constant parameter.
First of all, note that the expressions \eqref{Eqpesos1}-\eqref{Eqpesos2} indicate only the proportionality of the probabilities w.r.t.\ $t$ and $\red{\ell}$ or $\red{m}$, i.e., in order to obtain the probability mass function we have to normalize the weights above.
Moreover, observe that for $t > T_p$ the probability of choosing an index $\red{\ell}$ (resp. $\red{m}$) depends on the position $\red{\ell}$ (resp. $\red{m}$) and the time $t$.
More specifically, this probability increases with the value of $\red{\ell}$ (resp. $\red{m}$), and this effect grows as $t$ increases, with the probability mass function becoming a delta located at the last possible position when $t\rightarrow +\infty$.
{\red{The speed of convergence is controlled by the parameter $\beta$: the higher the value of $\beta$, the faster Eqs. \eqref{Eqpesos1} and \eqref{Eqpesos2} become delta functions.}}

\subsection{Cost functions: Bayesian risk minimization}
\label{sec:trainingProblem}


Let us define two matrices, $\mat{X} = [\x^{(1)}, \ldots, \x^{(N)}]$ and $\mat{Y} = [\y^{(1)}, \ldots, \y^{(N)}]$, containing all the features and observations in the training set respectively.
Furthermore, let us assume that the data associated with different training instances are independent, i.e.,
\begin{equation}
	\hat{p}(\mat{Y}|\mat{X},\s) = \prod_{n=1}^{N}{\hat{p}(\y^{(n)}|\x^{(n)},\s)}
		= \prod_{n=1}^{N}{\hat{p}(\y_{\s}^{(n)}|\x^{(n)})}.
\label{eq:condPdfData}
\end{equation}
From a Bayesian point of view, the best model (i.e., the best chain or label order) is the one that minimizes the \key{Bayesian risk} \cite{PCC,OnLabelDependenceJournal,VanTrees:detectionEstimationLinearModulation2001}.
{\pink{Let us define a generic cost function,
\begin{equation}
	\cost(\mat{Y},\hat{\mat{Y}}) = \mathcal{F}\left(\{\loss(\y_{\s}^{(n)},\yest_{\s}^{(n)})\}_{n=1}^{N}\right),
\label{eq:costGeneric}
\end{equation}
where we have used $\hat{\mat{Y}} = \hat{\mat{Y}}(\s) = \mat{H}(\mat{X}|\s)$ and $\yest_{\s}^{(n)} = \h_{\s}(\x^{(n)})$ to simplify the notation, $\mathcal{F}(\cdot)$ is a generic functional and $\loss(\y,\hat{\y})$ is some appropriate loss function, $\loss: \Y \mapsto \mathbb{R}$.}}
The Bayesian risk is the expected cost over the joint {\red{density}} of the data given the model,
\begin{equation}
	\risk(\s) = \expect{\vec{XY|s}} \{\cost(\mat{Y},\hat{\mat{Y}})\},
\label{eq:risk}
\end{equation}
with $\expect{\vec{XY|s}}$ denoting the mathematical expectation w.r.t.\ the joint conditional density $\hat{p}(\mat{X},\mat{Y}|\s)$, and the optimum chain corresponding to the label order which minimizes this risk.
%
For a given set of training data, the best label order can be determined in a pointwise way by taking the expectation w.r.t.\ the conditional probability \cite{PCC,OnLabelDependenceJournal}:\footnote{Note that in \cite{PCC,OnLabelDependenceJournal} this approach is followed to find the best classifier for a given label order, whereas here we use it to find the best label order (i.e., the best model).}
\begin{align}
	\hat{\s}(\mat{X}) & = \argmin_{\s \in \S^L}\ \expect{\vec{Y|X,s}} \{\cost(\mat{Y},\mat{H}(\mat{X}|\s))\} \nonumber \\
		& = \argmin_{\s \in \S^L} \sum_{\mat{Y} \in {\boldsymbol \Y}^N}{\cost(\mat{Y},\mat{H}(\mat{X}|\s))
			\prod_{n=1}^{N}{\hat{p}(\y_{\s}^{(n)}|\x^{(n)})}},
\label{eq:labelOrderCond}
\end{align}
where we have made use of \eqref{eq:condPdfData} to obtain the last expression.\footnote{In practice, we use internal validation to avoid overfitting: the training set is divided into two: a first part for training the classifiers and a second part for validation. Thus, all the expressions in this section should really consider only the validation set, which will be a subset of the training set. However, in the following we always consider $n=1,\ldots,N$ for the sake of simplicity.}
%

In the following, we explore several cost and loss functions commonly used in MLC and MDC, showing their probabilistic interpretation from the point of view of finding the best label order.

\subsubsection{Additive cost functions}
\label{sec:additiveCost}

In this section we consider the functional $\mathcal{F}(\cdot) = \sum_{n=1}^{N}(\cdot)$, i.e., an additive cost function. Thus, we have
\begin{equation}
	\cost_{\textrm{sum}}(\mat{Y},\hat{\mat{Y}}) = \sum_{n=1}^{N}{\loss(\y_{\s}^{(n)},\yest_{\s}^{(n)})}.
\label{eq:costSum}
\end{equation}
Inserting \eqref{eq:costSum} into \eqref{eq:labelOrderCond}, and after some algebra, we obtain the following estimator for additive cost functions:
%
\begin{equation}
	\hat{\s}(\mat{X}) = \argmin_{\s \in \S^L} \sum_{n=1}^{N} \sum_{\y_{\s}^{(n)} \in {\boldsymbol \Y}} 
		\loss(\y_{\s}^{(n)},\yest_{\s}^{(n)}) \hat{p}(\y_{\s}^{(n)}|\x^{(n)}).
\label{eq:labelOrderAdditive2}
\end{equation}

Unfortunately, minimizing \eqref{eq:labelOrderAdditive2} for a generic loss function can be unfeasible in practice.
However, by focusing on two of the most common losses used in MLC and MDC (the exact match and the Hamming losses), simple expressions with a straightforward probabilistic interpretation may be found.
First of all, let us consider the \emph{exact match} loss,\footnote{Also called by some authors the \emph{subset 0/1} loss (cf.~\cite{PCC}).} which is defined as
\begin{align}
	\loss_{\textrm{EM}}(\y^{(n)},\yest_{\s}^{(n)}) & = \llbracket \y_{\s}^{(n)} \ne \yest_{\s}^{(n)}) \rrbracket 
		 = \begin{cases}
				1, & \y_{\s}^{(n)} \ne \yest_{\s}^{(n)};\\
				0, & \y_{\s}^{(n)} = \yest_{\s}^{(n)};
			\end{cases}
\label{eq:exactMatchLoss}
\end{align}
where $\llbracket \cdot \rrbracket$ returns 1 if its predicate holds and 0 otherwise.
Using \eqref{eq:exactMatchLoss}, \eqref{eq:labelOrderAdditive2} can be expressed as
\begin{align}
	\hat{\s}_{\textrm{EM}}(\mat{X}) & = \argmin_{\s \in \S^L}
			\sum_{n=1}^{N}{\sum_{\y_{\s}^{(n)} \in {\boldsymbol \Y}}{\llbracket \y_{\s}^{(n)} \ne \yest_{\s}^{(n)} \rrbracket
				\hat{p}(\y_{\s}^{(n)}|\x^{(n)})}} \nonumber \\
		& = \argmin_{\s \in \S^L} \sum_{n=1}^{N}{\left(1-\hat{p}(\yest_{\s}^{(n)}|\x^{(n)})\right)} \nonumber \\
		& = \argmax_{\s \in \S^L} \sum_{n=1}^{N}{\hat{p}(\yest_{\s}^{(n)}|\x^{(n)})}.
\label{eq:labelOrderEM}
\end{align}
From \eqref{eq:labelOrderEM} it can be seen that minimizing the exact match loss is equivalent to maximizing the sum of the likelihoods of the predictions for each of the instances in the validation set.\footnote{\pink{Note that this is equivalent to the result obtained in \cite{PCC,OnLabelDependenceJournal} for the test stage, i.e., for inferring the best $\hat{\y}_{\s}$ for a given label order $\s$.}}
Therefore, in order to minimize the exact match loss we should use the following payoff function:
\begin{equation}
	J_{\textrm{EM}}(\s) = \sum_{n=1}^{N}{\hat{p}(\yest_{\s}^{(n)}|\x^{(n)})}.
\label{eq:payoffEM}
\end{equation}

As a second example, we consider the \key{Hamming} loss:\footnote{The name is due to the fact that it corresponds to the Hamming distance for the binary labels used in MLC. Although this is no longer true for the non-binary labels that can appear in MDC, this definition is still valid and we keep the name used in MLC.}
\begin{equation}
	\loss_{\textrm{Ham}}(\y_{\s}^{(n)},\yest_{\s}^{(n)})
		= \sum_{\ell=1}^{L}{\llbracket y_{s_{\l}}^{(n)} \ne \hat{y}_{s_{\l}}^{(n)} \rrbracket}.
\label{eq:hammingLoss}
\end{equation}
Unlike the exact match loss, which returns the same value when $y_{s_{\l}}^{(n)} \ne \hat{y}_{s_{\l}}^{(n)}$ regardless of how dissimilar they are, the Hamming loss looks at each label component separately.
Using \eqref{eq:hammingLoss}, it can be shown (see the Appendix) that, for the Hamming loss, \eqref{eq:labelOrderAdditive2} becomes
\begin{equation}
	\hat{\s}_{\textrm{Ham}}(\mat{X}) = \argmax_{\s \in \S^L}
		\sum_{n=1}^{N}{\sum_{\l=1}^{L}{\pink{\hat{p}(\hat{y}_{s_{\l}}^{(n)}|\x^{(n)})}}}.
\label{eq:labelOrderHamming}
\end{equation}
Hence, from \eqref{eq:labelOrderHamming} we notice that the Hamming loss is minimized by maximizing the sum of the likelihoods of the individual label predictions{\red{, given \emph{only} the data,}} for each of the instances in the validation set.\footnote{\pink{Once more this is equivalent to the result obtained in \cite{PCC,OnLabelDependenceJournal} for the test stage.}}$^{,}$\footnote{\pink{Note that the \texttt{CC} approach returns $\hat{p}(\hat{y}_{s_{\l}}^{(n)}|\x^{(n)}, \hat{y}_{s_{1}}^{(n)}, \ldots, \hat{y}_{s_{\l-1}}^{(n)})$ instead of $\hat{p}(\hat{y}_{s_{\l}}^{(n)}|\x^{(n)})$. However, an estimate of the probabilities required by \eqref{eq:labelOrderHamming} and \eqref{eq:payoffHamming} can be easily obtained by summing over the unnecessary variables, i.e.,
\begin{equation*}
	\hat{p}(\hat{y}_{s_{\l}}^{(n)}|\x^{(n)})
		= \sum_{\hat{y}_{s_{1}}^{(n)}, \ldots, \hat{y}_{s_{\l-1}}^{(n)}}
			{\hat{p}(\hat{y}_{s_{\l}}^{(n)}|\x^{(n)}, \hat{y}_{s_{1}}^{(n)}, \ldots, \hat{y}_{s_{\l-1}}^{(n)})}.
\end{equation*}}}
Thus, the corresponding payoff required for minimizing the Hamming loss is
\begin{equation}
	J_{\textrm{Ham}}(\s) = \sum_{n=1}^{N}{\sum_{\l=1}^{L}{\pink{\hat{p}(\hat{y}_{s_{\l}}^{(n)}|\x^{(n)})}}}.
\label{eq:payoffHamming}
\end{equation}

\subsubsection{Multiplicative cost functions}
\label{sec:multiplicativeCost}

As a second family of cost functions we consider multiplicative cost functions, i.e., we consider a functional $\mathcal{F}(\cdot) = \prod_{n=1}^{N}{(\cdot)}$, which leads us to
\begin{equation}
	\cost_{\textrm{prod}}(\mat{Y},\hat{\mat{Y}}) = \prod_{n=1}^{N}{\loss(\y_{\s}^{(n)},\yest_{\s}^{(n)})}.
\label{eq:costProd}
\end{equation}
Inserting \eqref{eq:costProd} into \eqref{eq:labelOrderCond}, the estimator is now given by
\begin{align}
	\hat{\s}(\mat{X}) & = \argmin_{\s \in \S^L} \sum_{\mat{Y} \in {\boldsymbol \Y}^N}
			{\prod_{n=1}^{N}{\loss(\y_{\s}^{(n)},\yest_{\s}^{(n)}) \hat{p}(\y_{\s}^{(n)}|\x^{(n)})}} \nonumber \\
		& = \argmin_{\s \in \S^L} \prod_{n=1}^{N}{\sum_{\y_{\s}^{(n)} \in {\boldsymbol \Y}}{\loss(\y_{\s}^{(n)},\yest_{\s}^{(n)}) \hat{p}(\y_{\s}^{(n)}|\x^{(n)})}} \nonumber \\
		& {\pink{= \argmin_{\s \in \S^L} \sum_{n=1}^{N}{\log\left(\sum_{\y_{\s}^{(n)} \in {\boldsymbol \Y}}{\loss(\y_{\s}^{(n)},\yest_{\s}^{(n)}) \hat{p}(\y_{\s}^{(n)}|\x^{(n)})}\right)}}},
\label{eq:labelOrderMult}
\end{align}
{\pink{which has an similar functional form to \eqref{eq:labelOrderAdditive2}, with the $\log$ of the inner sum inside the outer sum.
Hence, following an identical procedure to the one in Eq. \eqref{eq:labelOrderEM} for the exact match loss, we obtain}}
\begin{align}
	\hat{\s}_{\textrm{EM-prod}}(\mat{X}) = & \argmin_{\s \in \S^L} {\pink{\sum_{n=1}^{N}{\log\left(1-\hat{p}(\yest_{\s}^{(n)}|\x^{(n)})\right)}}} \nonumber \\
		= & \argmax_{\s \in \S^L} \prod_{n=1}^{N}{\hat{p}(\yest_{\s}^{(n)}|\x^{(n)})},
\label{eq:labelOrderMultEM}
\end{align}
which corresponds to the maximum of the likelihood function.
Hence, the corresponding payoff function is precisely the likelihood function:
\begin{equation}
	J_{\textrm{EM-prod}}(\s) = \prod_{n=1}^{N}{\hat{p}(\yest_{\s}^{(n)}|\x^{(n)})},
\label{eq:payoffEMprod}
\end{equation}
{\pink{Similarly, following the steps shown in the Appendix for the additive cost function, we may obtain the estimator for the Hamming loss in the multiplicative case:}}
\begin{equation}
	\hat{\s}_{\textrm{Ham-prod}}(\mat{X}) = \argmax_{\s \in \S^L} \prod_{n=1}^{N}
		{\sum_{\l=1}^{L}{{\pink{\hat{p}(\hat{y}_{s_{\l}}^{(n)}|\x^{(n)})}}}}.
\label{eq:labelOrderMultHamming}
\end{equation}
which is similar to \eqref{eq:labelOrderMultEM}, but now the product is on the individual label likelihoods instead of the global likelihoods of the different instances.
The payoff function in this case is
\begin{equation}
	J_{\textrm{Ham-prod}}(\s) = \prod_{n=1}^{N}{\sum_{\l=1}^{L}{{\pink{\hat{p}(\hat{y}_{s_{\l}}^{(n)}|\x^{(n)})}}}}.
\label{eq:payoffHammingProd}
\end{equation}

\section{Test (inference) stage: Finding the best label vector}
\label{sec:test}

In the test stage, for a given test instance $\xtest$ and a label order $\s$, our aim is finding the optimal label vector $\hat{\y}_{\s}$ that maximizes \Eq{eq:pcc}.
The \texttt{PCC} method \cite{PCC} solves this part analytically (by performing an exhaustive search). However, since this method becomes computationally intractable for anything but small $L$ (the full space involves $\prod_{\l=1}^{L} K_{\l}$ possible paths).

The goal is providing a Monte Carlo (MC) approximation of the estimated label vector,
\begin{equation}
	{\bf \hat{y}}_{\s}^{({\bf mc})} \approx {\bf \hat{y}}_{\s} = \argmax_{\y_{\s}}\ \hat{p}(\y_{\s} | \xtest)
\end{equation}
{\pink{for the minimization of the exact-match loss or
\begin{equation}
	{\bf \hat{y}}_{\s}^{({\bf mc})} \approx {\bf \hat{y}}_{\s} = \argmax_{\y_{\s}}\ \sum_{\ell=1}^{L}{\hat{p}(y_{s_{\ell}} | \xtest)}
\end{equation}
for the Hamming loss,}} such that ${\bf \hat{y}}_{\s}^{({\bf mc})} \rightarrow {\bf \hat{y}}_{\s}$ when $T_y \rightarrow +\infty$, with $T_y$ being the number of iterations of the MC algorithm.

A first  possible MC approach {\pink{for the minimization of the exact match loss}} is provided by \Code{alg:mcmc}.\footnote{\pink{\Code{alg:mcmc} can also be used to minimize the Hamming loss, simply changing the condition in step 1(a) by the following condition:
\begin{equation*}
	\sum_{\ell=1}^{L}{\hat{p}(y_{s_{\ell}}' | \xtest)} > \sum_{\ell=1}^{L}{\hat{p}(y_{s_{\ell}}^{(t-1)} | \xtest)}.
\end{equation*}}}$^,$\footnote{\pink{An MC-based approach like the one shown in \Code{alg:mcmc} has been independently proposed in \cite{ChainingMLC} for the minimization of the exact match loss during the test stage.}}
Given a test instance $\x^*$ and a label order $\s$, this algorithm starts from an initial label vector $\y^{(0)}_{\s}$ arbitrarily chosen (e.g., randomly or from the greedy inference offered by standard \texttt{CC}), and draws samples $\y_{\s}^{(i)}$ ($i=1,\ldots,T_y$) directly from the model learnt in the training stage, $\hat{p}(\y_{\s}|\xtest)$.\footnote{Note that this is equivalent to generating random paths in the tree of class labels according to the corresponding weights associated to each branch (see Figure \ref{Fig3}).}
Then, the label vector $\y_{\s}^{(k)}$ with the highest {\red{payoff}} is returned as the output, i.e., ${\bf \hat{y}}_{\s}^{({\bf mc})} =\y_{\s}^{(k)}$, with
\begin{equation}
	k = \argmax_{i=1,\ldots,T_y}\ \hat{p}(\y_{\s}^{(i)}|\xtest)
\end{equation}
{\pink{for the minimization of the exact-match loss and
\begin{equation}
	k = \argmax_{i=1,\ldots,T_y}\ \sum_{\ell=1}^{L}{\hat{p}(y_{s_{\ell}}^{(i)}|\xtest)}
\end{equation}
when the goal is minimizing the Hamming loss.}}
From a Monte Carlo point of view, it is important to remark that all the candidate vectors $\y'$ are always drawn {\it directly} from the target density, $\hat{p}(\y_{\s}|\xtest)$, i.e., $\y'$ is always a valid path on a tree selected according to the weights of the different branches.
This is an important consideration, since it guarantees that the estimated label vector, ${\bf \hat{y}}_{\s}^{({\bf mc})}$, will always be a feasible path.

\begin{algorithm}[!tb]
\caption{\label{alg:mcmc} Obtaining ${\bf \hat{y}}_{\s}^{({\bf mc})}\approx\hat{\y}_{\s}$ {\red{that minimizes the exact-match loss}} for a given test instance $\xtest$.}
Input:
\begin{itemize}[noitemsep,nolistsep]
	\item $\xtest, \s$: test instance and given label order.
  \item $\hat{p}(\y_{\s}|\x)$: probabilistic model.
  \item $\y_{\s}^{(0)},T_y$: initial label vector and number of iterations.
\end{itemize}
Algorithm:
\begin{enumerate}[noitemsep,nolistsep]
	\item For $t=1,\ldots,T_y$:
	\begin{enumerate}[noitemsep,nolistsep]
		\item Draw $\y_{\s}' \sim \hat{p}(\y_{\s}|\xtest)$.
		\item \texttt{if} $\hat{p}(\y_{\s}'|\xtest) > \hat{p}(\y_{\s}^{(t-1)}|\xtest)$
		\begin{itemize}[noitemsep,nolistsep]
			\item $\y_{\s}^{(t)} \gets \y_{\s}'$ accept.
		\end{itemize}
		\item \label{accept2} \texttt{else} 
		\begin{itemize}[noitemsep,nolistsep]
			\item $ \y_{\s}^{(t)} \gets\y_{\s}^{(t-1)}$ reject.
		\end{itemize}
	\end{enumerate}
\end{enumerate}
	Output:
	\begin{itemize}[noitemsep,nolistsep]
		\item ${\bf \hat{y}}_{\s}^{({\bf mc})}  = \y_{\s}^{(T_y)}$: predicted label assignment.
	\end{itemize}
\end{algorithm}

As previously discussed, the inference of \Code{alg:mcmc} depends strictly on the chosen label order $\s$.
For this reason, we also propose another scheme that uses a population of label orders $\mat{S}=[\s^{(1)},\ldots,\s^{(M)}]$ (chosen randomly or obtained using \Code{alg:full2}).
A naive procedure to incorporate this information in the inference technique would be running $M$ parallel algorithms to find sequences of labels ${\y}_{\s}^{(i)}$ (like \Code{alg:mcmc}) using different label orders $\spred^{(i)}$ ($i=1,\ldots,M$) and then selecting the best one.
However, this approach is computationally inefficient.
In \Code{alg:mcmc2} we propose a more sophisticated approach that makes use of the information within the entire population $\mat{S}$ but requires running only one random search.
The main steps of the method can be summarized as follows: 
\begin{enumerate}
	\item A label order $\s' \in \mat{S}$ is selected according to some weights (e.g., those provided by \Code{alg:full2}) proportional to a certain payoff
		function.
	\item A good label vector $\hat{\y}$ is found by following \Code{alg:mcmc}.
	\item The procedure is repeated $T_y$ times, with the best label vector for the $T_s$ label orders explored being returned as the output.
\end{enumerate}

\begin{algorithm}
\caption{\label{alg:mcmc2} Obtaining ${\bf \hat{y}}_{\s}^{({\bf mc})}\approx\hat{\y}_{\s}$ {\red{that minimizes the exact-match loss}} given $\xtest$, and a population $\mat{S}$.}
Input:
\begin{itemize}[noitemsep,nolistsep]
	\item $\xtest$: test instance.
	\item $\mat{S}=[\s^{(1)},\ldots,\s^{(M)}]$: population of $M$ label orders.
	\item $w^{(1)},\ldots,w^{(M)}$: corresponding weights.
  \item $\hat{p}(\y_{\s}|\x) = \hat{p}(\y|\x,\s)$: probabilistic model.
  \item $T_s,T_y$: number of iterations for searching $\s$ and $\y_{\s}$ resp.
  \item $\y_{\s}^{(0)}$:  initial label vector.
\end{itemize}
Algorithm:
\begin{enumerate}[noitemsep,nolistsep]
	\item For $t_1=1,\ldots,T_s$:
	\begin{enumerate}[noitemsep,nolistsep]
		\item Choose $\s_{t_1} = s^{(j)} \sim w^{(j)}/\sum_{i=1}^{M}{w^{(i)}}$ for $j=1,\ldots,M$.
		\item Set $\z_{1}=\y_{\s_{t_1-1}}^{(t_1-1)}$. 
		\item For $t_2=1,\ldots,T_y$:
		\begin{enumerate}[noitemsep,nolistsep]
    	\item $\z' \sim p(\z|\xtest,\s_{t_1})$.
	    \item \texttt{if} $\hat{p}(\z'|\xtest,\s_{t_1}) > \hat{p}(\z_{t_2}|\xtest, \s_{t_1})$
		  \begin{itemize}[noitemsep,nolistsep]
				\item $\z_{t_2+1}\gets \z'$ accept.
	    \end{itemize}
	    \item \label{accept1} \texttt{else} 
	    \begin{itemize}[noitemsep,nolistsep]
		  	\item $\z_{t_2+1} \gets \z_{t_2}$ reject.
	    \end{itemize}
		\end{enumerate}
	\item Set $\y_{\s_{t_1}}^{(t_1)}=\z_{T_y}$.
	\end{enumerate}
\end{enumerate}
Output:
\begin{itemize}[noitemsep,nolistsep]
	\item ${\bf \hat{y}}_{\s}^{({\bf mc})}=\y_{\s_{T_s}}^{(T_s)}$: predicted label assignment.
\end{itemize}
\end{algorithm}

\section{Experiments}
\label{sec:Exp}

In order to compare fairly both the performance and the computational effort, we progressively apply the ideas introduced in the previous sections to form four novel methods:
%
%
%
\begin{itemize}
	\item \texttt{MCC} (\Code{alg:mcmc}): given a classifier chain trained on some previously-determined label order $\s$ (e.g., randomly as in \texttt{CC} or \texttt{PCC}),
		we infer the label vector for all test instances using a simple MC approach.
	\item \texttt{M$_\s$CC} (\Code{alg:full} plus \Code{alg:mcmc}): like \texttt{MCC}, but we additionally search for a suitable label order $\spred$ during the training
		stage. Specifically, we use \Code{alg:full} with the simplest proposal density $\pi(\s|\s_{t-1})$ described in the first part of Section \ref{sec:proposal}. 
	\item \texttt{PM$_\s$CC} (\Code{alg:full2} plus \Code{alg:mcmc2}): population version of \texttt{M$_\s$CC}, still using the simplest proposal density $\pi(\s|\s_{t-1})$
		described in Section \ref{sec:proposal}.
	\item \texttt{P$_t$M$_\s$CC} (\Code{alg:full2} plus \Code{alg:mcmc2}): \texttt{PM$_\s$CC} with the improved proposal $\pi(\s|\s_{t-1})$ described in the last part of Section \ref{sec:proposal}.
\end{itemize}
Note that  \texttt{MCC} and  \texttt{M$_\s$CC} differ on how they obtain the label order $\s$ (randomly chosen for \texttt{MCC} or estimated using \Code{alg:full} for \texttt{M$_\s$CC}), whereas \texttt{PM$_\s$CC} and \texttt{P$_t$M$_\s$CC} differ on the proposal used to search for the best label order.

\subsection{Comparison of different cost functions}
\label{sec:cost}

In this section we analyze the performance of different payoff functions: the two additive payoffs given by \eqref{eq:payoffEM} and \eqref{eq:payoffHamming}, and the multiplicative payoff of \Eq{eq:payoffEMprod}.
Initially we focus on the \textsf{Music} dataset, because it is faster to run and easier to visualise than other datasets.
Indeed, since $L=6$ (see Table \ref{table:datasets}) we can find the optimum label order for the exact match payoff ($\spred_{\textrm{EM}}=[3,5,0,1,4,2]^{\top}$) by performing an exhaustive search over the $L!=720$ possibilities.
\Tab{fig:1} shows that the proposed Monte Carlo approach (\Code{alg:full}) arrives to the optimum label order under two separate initializations after 1935 and 1626 iterations respectively; although we note that after a much smaller number of iterations (310 and 225 respectively), the difference is minimal in payoff.
The search also converges maximizing $\Jep$ (Table not displayed), although we noted that it is a \emph{different} maxima, specifically, $\spred_{\textrm{EM-prod}}=[4,5,1,2,3,0]^{\top}$.

\begin{table}
\centering
\caption{\label{fig:1}Running \Code{alg:full} on \textsf{Music} dataset using payoff $\Je$ under $T_s=\infty$ (i.e., run until convergence to the optimum label order, $\spred_{\textrm{EM}}=[3,5,0,1,4,2]^{\top}$, obtained through an exhaustive search). We only show the iterations where a new $\s_t \gets \s'$ is accepted (plus the default $\s_0$); displaying also the payoffs $\Je$, $\Jep$ (in the log domain) and $\Jh$ (note that these numbers have \emph{not} been normalized by $N$). The experiment is performed twice for two different random seeds (i.e., starting from a different  $\s_0$).}
	\footnotesize
	\subfloat[a][Random seed $1$]{
		\begin{tabular}{llllll}
		\hline
			$t$ & $\s_t^{\top}$  & $\Je$ & $\Jep$ & $\Jh$ \\
		\hline
		$0$ & [5, 2, 4, 1, 0, 3]&    164.92& -1079.26&  2889.89\\
		$2$ & [5, 4, 0, 1, 2, 3]&    166.14& -1084.91&  2887.3\\
		$4$ & [5, 3, 0, 2, 1, 4]&    166.89& -1085.4 &  2886.84\\
		$310$ & [5, 3, 0, 1, 4, 2]&    167.42& -1084.58&  2887.14\\
		$492$ & [3, 5, 1, 0, 4, 2]&    167.53& -1083.97&  2887.41\\
		$1682$ & [3, 0, 5, 1, 4, 2]&    167.62& -1082.89&  2887.94\\
		$1935$ & [3, 5, 0, 1, 4, 2]&    167.73& -1082.79&  2887.91\\
		\hline
		\end{tabular}
	}

	\subfloat[a][Random seed $2$]{
		\begin{tabular}{llllll}
		\hline
			$t$ & $\s_t^{\top}$  & $\Je$ & $\Jep$ & $\Jh$ \\
		\hline
		$0$ & [4, 2, 0, 1, 3, 5]&    155.7 & -1109.58&  2864.41\\
		$1$ & [4, 2, 0, 3, 1, 5]&    156.87& -1102.82&  2868.08\\
		$2$ & [4, 2, 0, 3, 5, 1]&    159.45& -1096.79&  2873.95\\
		$3$ & [4, 0, 2, 3, 5, 1]&    161.79& -1091.6 &  2880.18\\
		$5$ & [4, 0, 5, 2, 3, 1]&    163.14& -1093.32&  2880.16\\
		$18$ & [5, 1, 4, 3, 2, 0]&    163.59& -1085.69&  2885.91\\
		$23$ & [5, 4, 0, 1, 2, 3]&    166.12& -1084.97&  2887.24\\
		$128$ & [3, 5, 1, 0, 2, 4]&   167.05& -1084.7 &  2887.2\\
		$176$ & [5, 3, 1, 0, 4, 2]&    167.22& -1085.75&  2886.65\\
		$225$ & [5, 3, 1, 4, 0, 2]&    167.41& -1083.1 &  2887.93\\
		$1422$ & [3, 5, 1, 4, 0, 2]&    167.69& -1081.35&  2888.68\\
		$1626$ & [3, 5, 0, 1, 4, 2]&    167.73& -1082.82&  2887.9\\
		\hline
		\end{tabular}
	}
\end{table}

\begin{figure}
    \centering 
	\subfloat[a][Random seed $1$]{\includegraphics[width=0.49\textwidth]{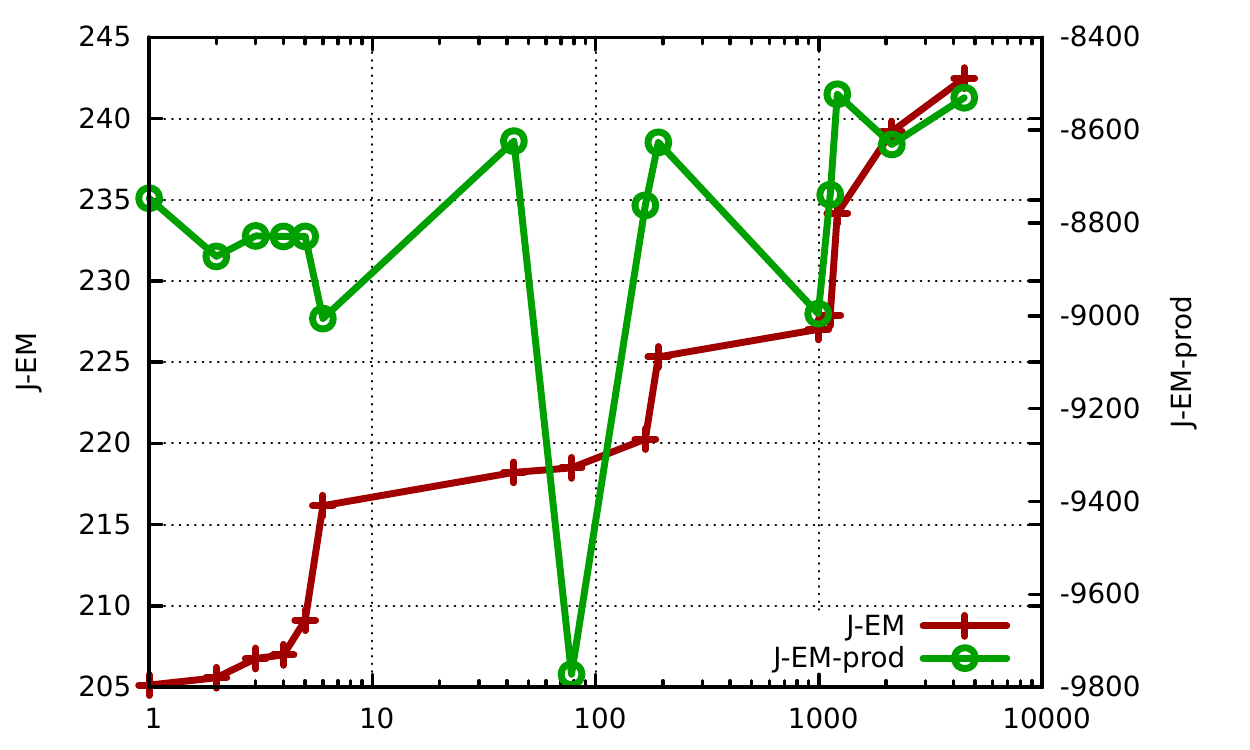}}
	\subfloat[b][Random seed $2$]{\includegraphics[width=0.49\textwidth]{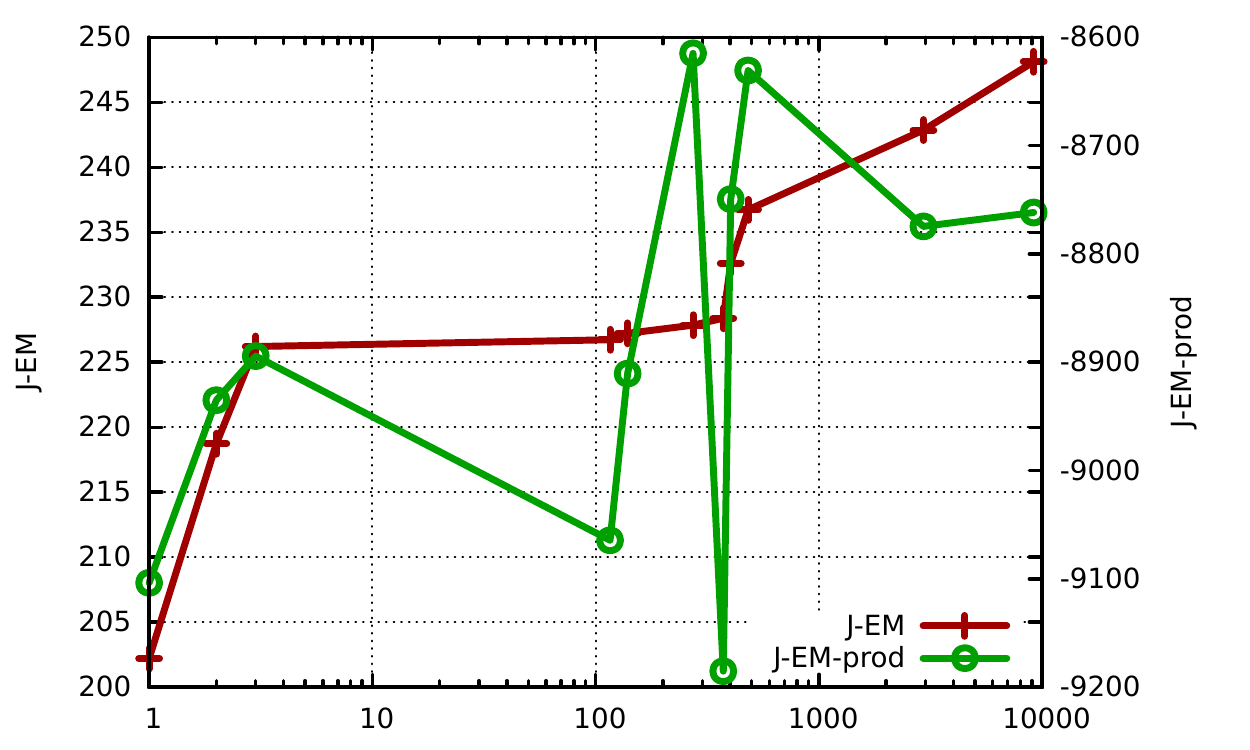}}
	\caption{\label{figure:y} A plotted version of \Tab{fig:1} for the \textsf{Yeast} data, for both $\Je$ and $\Jep$  up till $T_s=10,000$.
		The left vertical axis corresponds to $\Je$, and the right vertical axis to $\Jep$. Note the log-scale horizontal axis.}
\end{figure}

\begin{figure}
    \centering 
	\subfloat[a][Maximizing $\Jep$]{\includegraphics[width=0.49\textwidth]{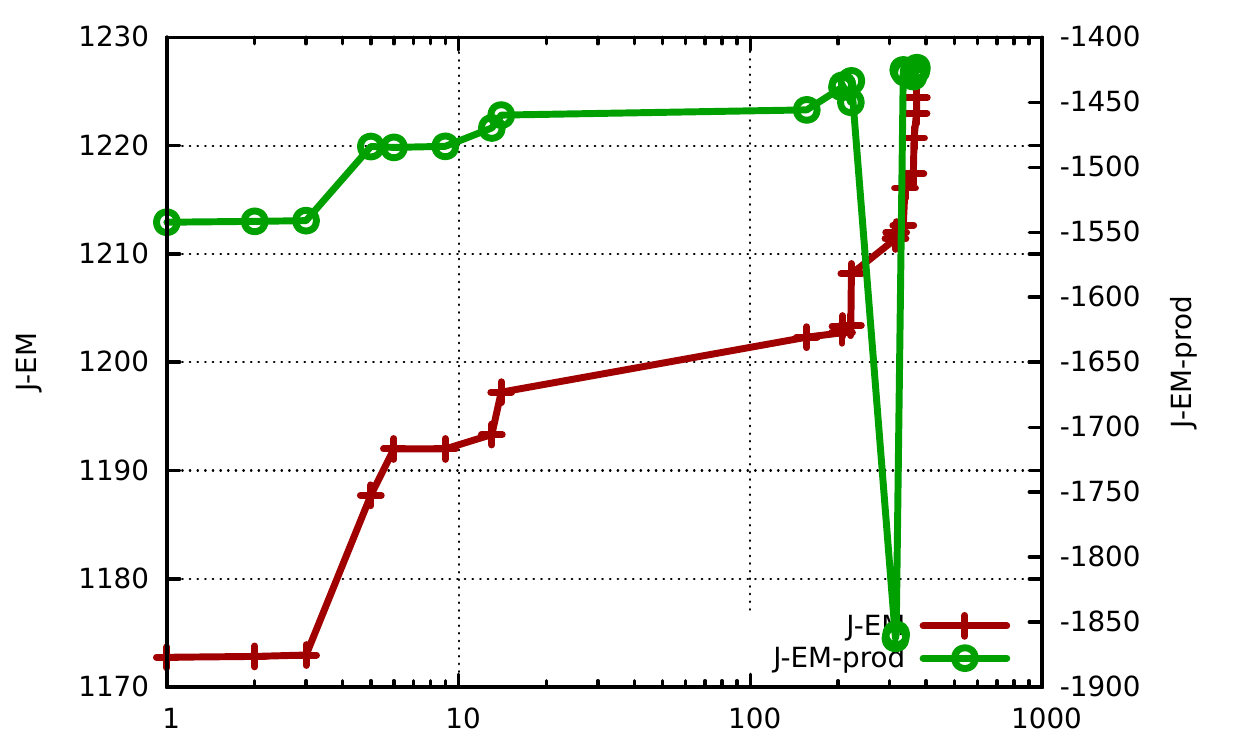}}
	\subfloat[a][Maximizing $\Je$]{\includegraphics[width=0.49\textwidth]{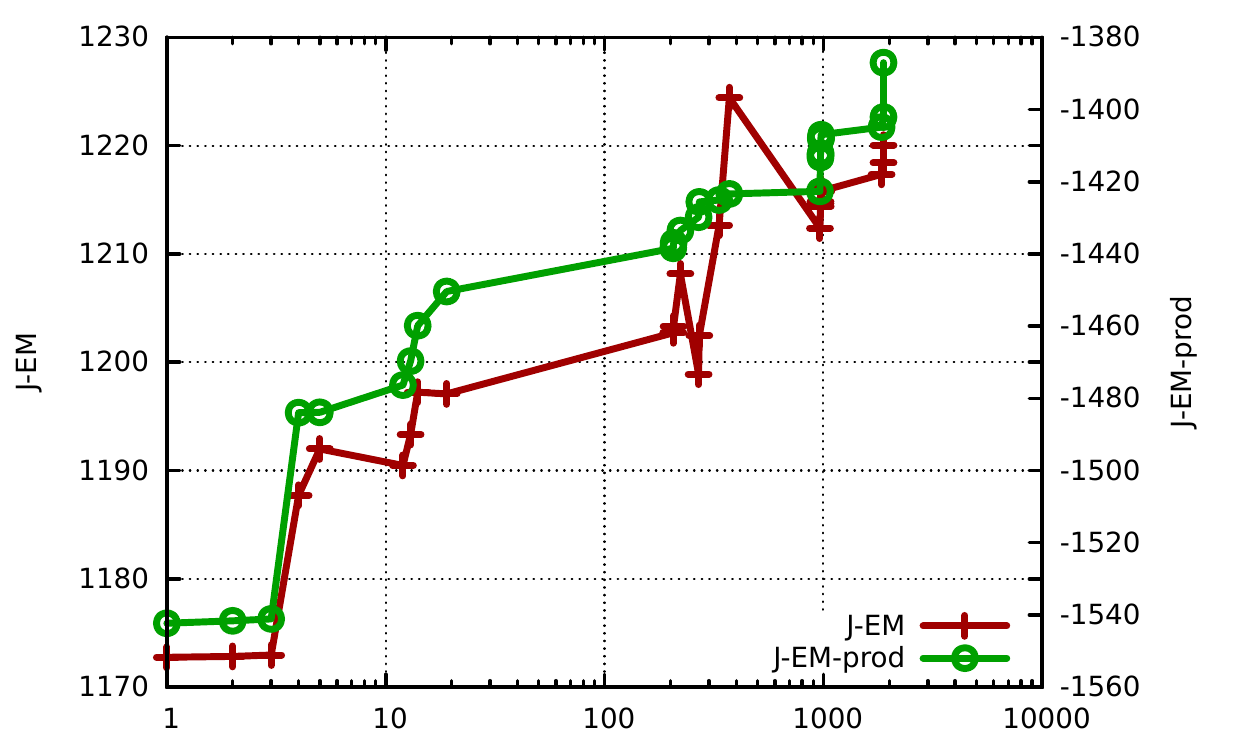}}
	\caption{\label{figure:e}Searching $\s$ space on the \textsf{Enron} dataset, displaying \emph{both} payoff scores, but maximizing only \emph{one} of them
		($\Je$ left, $\Jep$ right) -- using the same random seed (initial $\s_0$); until $T_s=10,000$. The left vertical axis corresponds to $\Je$, and the right
		vertical axis to $\Jep$. Note the log-scale horizontal axis.}
\end{figure}

A similar analysis may be performed for other datasets where the optimum label order cannot be found by exhaustive search.
\Fig{figure:y} plots similar statistics for the \textsf{Yeast} data ($L=14$), whereas \Fig{figure:e} shows the payoffs when maximizing them \emph{separately} (using the same random seed in both graphs) for \textsf{Enron} ($L=53$).
All these analysis suggest that the payoff functions are climbing the same terrain, but there are many peaks of similar height.
Thus, while $\spred = [4,5,1,2,3,0]^{\top}$ does not appear superficially close to $\spred = [3,5,0,1,4,2]^{\top}$, as found by the using the different payoff functions on \textsf{Music}, both result in higher performance than selecting $\s$ randomly.
This is also confirmed by the results of predictive performance shown later on, and justifies searching the $\S$-space.
The fact that many label orders provide good results, as opposed to just one, is not unexpected and justifies our population Monte Carlo method (\Code{alg:full2}).
As a general remark, we also note that the terrain of $\Jep$ appears much rougher: when maximizing $\Jep$, $\Je$'s appreciation is still relatively smooth, but not vice versa.

\Tab{table:1} compares the predictive performance using different payoff functions.
$\Je$ performs better than $\Jep$, which has a rougher terrain to climb.
However, we remark again that all of them are better than choosing a random $\s$.
This corresponds with our intuition and theoretical results, although perhaps even more experimentation is necessary to resolve the question in the formal statistical significant sense; due to randomly varying the $\D$-split and the initial $\s_0$, the payoff function, plus taking into account the huge $\S$ space, a vast number of experiments would be necessary for getting conclusive statistical-significant figures.
We instead decided to invest more computation in our large-scale comparison between methods (\Sec{sec:Results}).

\begin{table*}
\centering
\caption{\label{table:1}Average exact match and Hamming score across $100$ experiments (2/3 : 1/5 random data split), under different payoff functions, including \emph{none} (i.e., random $\s$).}
\footnotesize
	\subfloat[a][\textsf{Music}, $T_s = 1,500$]{
		\begin{tabular}{llllll}
		\hline
			Payoff & Exact Match\\
		\hline
		$\Je$     &     $  0.3386 \pm	 0.026$ \\
		$\Jep$     &      $ 0.3307  \pm    0.025$ \\
		$\Jh$     &       $0.3319  \pm   0.024$ \\
		None       &     $0.3264 \pm 0.022$ \\
		\hline
		\end{tabular}
		\begin{tabular}{llllll}
		\hline
			Payoff & Ham.\ Score\\
		\hline
			$\Je$     &       $0.7998  \pm  0.011$ \\
			$\Jep $    &       $0.7982 \pm  0.010$ \\
			$\Jh  $   &       $0.7988 \pm  0.010$ \\
	       None  &     $0.7935 \pm 0.014$ \\
		\hline
		\end{tabular}
	}\\
	\subfloat[a][\textsf{Yeast}, $T_s=100$]{
		\begin{tabular}{llllll}
		\hline
			Payoff & Exact Match\\
		\hline
		$\Je$     &     $ 0.2107    \pm   0.0126$\\
		$\Jep$     &     $ 0.2071	\pm  0.0108$\\
		$\Jh$     &     $ 0.2104	\pm  0.0115$\\
        None &     $ 0.2055	\pm  0.0115$\\
		\hline
		\end{tabular}
		\begin{tabular}{llllll}
		\hline
			Payoff & Ham. Score\\
		\hline
			$\Je$     & $	0.7815 \pm 0.0047$\\
			$\Jep $    & $	0.7816 \pm 0.0043$\\
			$\Jh  $   & $	0.7804 \pm 0.0046$\\
		     None   &     $0.7802  \pm 0.0048$\\
		\hline
		\end{tabular}
	}
\end{table*}

\subsection{Comparison of MCC with other MLC approaches}
\label{sec:compare}

\Tab{table:methods} outlines all methods we use in experiments, their parameters, and relevant references. We compare to baseline \texttt{IC}, the original classifier chains method \texttt{CC}, the ensemble version \texttt{ECC}, the Bayes-optimal rendition \texttt{PCC} and the $\epsilon$-approximate and beam search variants; the conditional dependency networks method \texttt{CDN}; and also \texttt{RF-PCT}, a decision-tree based approach.  

%
%
%
%

\begin{table*}[t]
    \centering 
    \caption{\label{table:methods}The methods considered and their parameters. The novel methods proposed are below the middle line; where each inherits the parameters of the previous ones, e.g., \texttt{P$_t$M$_\s$CC} takes parameters $T_y=100,T_s=50,M=10,\beta=0.03$. For \texttt{CDN}, $T_c$ is the number of collection iterations.
    	$M$ indicates generally the number of models. We selected the `width' parameter configuration for the Beam Search that gave the best \textsc{$0/1$ Loss} rank in \cite{BeamSearch}, and similarly for $\epsilon$-approximate PCC in \cite{ChainingMLC}. \texttt{RF-PCT} parameters are as in \cite{ExtML}}
	\begin{tabular}{llll}
		\toprule
		Key & Method & Parameters & Reference \\
		\midrule
		\texttt{IC} & Independent Classifiers&  & \cite{Overview}   \\
		\texttt{CC} & Classifier Chains  & & \cite{ECC2}  \\
		\texttt{ECC} & Ensembles of Classifier Chains & $M=10$ & \cite{ECC2}  \\
		\texttt{PCC} & Probabilistic Classifier Chains &  & \cite{PCC}  \\
		\texttt{CDN} & Conditional Dependency Networks & $T=1000$ & \cite{GuoGu}  \\
		             &                                 & $T_c=100$ &               \\
		\pink \texttt{PCC-$\epsilon$} & \pink $\epsilon$-approx.\ PCC & \pink $\epsilon=0.25$ & \cite{ChainingMLC}  \\
		\pink \texttt{PCC-Beam} & \pink Beam Search PCC & \pink $w_s^y= ^5_1$ & \cite{BeamSearch}  \\
		\pink \texttt{RF-PCT} & \pink Random Forest of PCTs & $M=100$ & \cite{RFPCT}  \\
		\midrule
		\texttt{MCC} & Monte Carlo Optimization for CC  & $T_y=100$ & Alg.~\ref{alg:full} \\
		\texttt{M$_\s$CC} & $2 \times$ \pink MCC  & $T_s=50$ & Algs.~\ref{alg:full}~$\&$~\ref{alg:mcmc} \\
		\texttt{PM$_\s$CC} & Population \texttt{M$_\s$CC} & $M=10$ & Algs.~\ref{alg:full2}~$\&$~\ref{alg:mcmc2} \\
		\texttt{P$_t$M$_\s$CC} & Population \texttt{M$_\s$CC} (w/ $t$emperature) & $\beta = 0.03$ & Algs.~\ref{alg:full2}~$\&$~\ref{alg:mcmc2} \\
		\bottomrule
	\end{tabular}
\end{table*}

As a base classifier (for all methods relying on one) we mainly use support vector machines (SVMs) fitted with logistic models (as according to \cite{SMOM}) so as to provide probabilistic output {\pink and otherwise with the default parameters as in the \texttt{SMO} implementation of the Weka framework \cite{WEKA}.}
Logistic regression has so far been a popular choice in the probabilistic multi-label literature (e.g., \cite{PCC,GuoGu}) due to its probabilistic output. However, we have found that SVM-based methods can perform better, {\pink at least without tuning the parameters. For best accuracy, it is highly recommended to tune the base classifier. However, we wish to avoid this ``dimension'' and instead focus on the multi-label methods.}
%
%

All our methods are implemented and will be made available within the Meka framework\footnote{\url{http://meka.sourceforge.net}}; an open-source framework based on the Weka machine learning framework \cite{WEKA} with added support for multi-label classification and evaluation. 

\Tab{table:datasets} displays the collection of real world datasets that we use; most are familiar to the MLC and MDC literature \cite{RAKEL, ECC2, PCC}.

\begin{table*}[t]
    \centering 
    \caption{\label{table:datasets}A collection of datasets and associated statistics, where \texttt{LC} is \key{label cardinality}: the average number of labels relevant to each example; relevant for binary labels \cite{Overview}. We have divided multi-dimensional datasets, and multi-label (binary-only) datasets.}
	\begin{tabular}{rrrrcccr}
	\toprule
	    		& $N$  &$L$ &	$K$ &$d$		&\texttt{LC} &Type 		 \\           
	\midrule           
	\data{Solar Flare}            &323   	&3      &5     &10 & N/A & astrology      \\
	\data{Bridges}                &107    	&5      &2--6  &7 & N/A & civil engineering  	   \\
	\data{Thyroid}                &9172   	&7      &2--5  &28 & N/A & medical  	\\
	\data{Parkinson's}            &488    	&5      &3     &58 & N/A & medical  	\\
	\midrule           
	\data{Music}       &593    &6 & 2      &72      &1.87          &audio   \\    
	\data{Scene}   	&2407	&6 & 2      &294		&1.07          &image	 \\     
	\data{Yeast}   	&2417 	&14 & 2  	&103	   	&4.24          &biology \\     
	\data{Genbase}     &661    &27 & 2     &1185    &1.25          &biology \\    
	\data{Medical}		&978 	&45 & 2     &1449    &1.25          &medical/text	 \\     
	\data{Enron}   	&1702	&53 & 2     &1001    &3.38          &email/text	 \\     
	\data{Reuters} 	&6000	&103 & 2    &500		&1.46          &news/text	 \\     
	\pink \data{TMC2007}    	&28596  &22   & 2   &500     &2.16 &text	 \\     
\pink 	\data{MediaMill}	&43907  &101    & 2 &120		&4.38	       &video	 \\     
	\bottomrule
	\end{tabular}
\end{table*}


%
{\pink
The two contrasting measures \eval{exact match loss} (\Eq{eq:exactMatchLoss}) and \eval{Hamming loss} (\Eq{eq:hammingLoss}) are almost invariably used in the multi-label literature, so we use them here. Note that in some results we pose both as a payoff/score $1 - \L$, where $\L$ denotes normalized loss; in other words, \eval{exact match} and \eval{Hamming score}, where $1.0$ is the best possible performance.
}

\subsection{Results}
\label{sec:Results}

\Tab{table:accuracy} displays the average results of $5$-fold cross validation. Results for running time performance are given in \Tab{table:rtime}. {\pink All experiments were carried out on Intel Xeon CPUs at 3.16GHz allowing up to 2 GB of RAM.} The ranks and average ranks of each method are displayed, and significance according to the Nemenyi test \cite{Nemenyi}; where $a \succ b$ indicates that algorithm $a$ is significantly better than $b$ (under a $p$-value of $0.10$). 

In \Tab{table:new}, to compare with existing methods from the literature, we have taken results from \cite{ChainingMLC,BeamSearch,ExtML}, and displayed results for our methods alongside using the same train/test splits\footnote{As made available on the Mulan website: \url{http://mulan.sourceforge.net/datasets.html}}. Note that the PCC methods use logistic regression as a base classifier, and \texttt{RF-PCT} is decision-tree based.

\begin{table*}
	\footnotesize
	\centering
	\pink
	\caption{\label{table:new}\pink Comparison of other methods from the literature on the train/test splits used in these papers. N/A indicates that the result is Not Available (the dataset was not used by the algorithm's authors). DNF indicates Did Not Finish in 24 hours. Ranks are not shown due to many missing values. Best results are highlighted in \textbf{bold}.}
	\textsc{$0/1$ Exact-Match Loss}\\
	\begin{tabular}{lllllll}
		\toprule
			           & \texttt{PCC-$\epsilon$} & \texttt{PCC-Beam}   & \texttt{RF-PCT} & \texttt{MCC} & \texttt{M$_\s$CC} & \texttt{P$_t$M$_\s$CC} \\
		\midrule              
            \data{Music}       &  0.718         & 0.673 & 0.693  &     0.673    & 0.688    & \textbf{0.653} \\
            \data{Scene}       &  0.385         & 0.362 & 0.482  &  0.419    & 0.382    & \textbf{0.360}    \\ 
            \data{Yeast}       &  0.764         & \textbf{0.758}  & 0.848  &     0.775    & 0.776    & 0.776    \\
            \data{Genbase}       &  N/A           & \textbf{0.020} & N/A    & {\bf 0.020}    & {\bf 0.020}    & {\bf 0.020} \\
            \data{Medical}    &  0.541         & \textbf{0.360} & 0.372  & 0.364     & {\bf 0.360}    &  {\bf 0.360}    \\ 
            \data{Enron}        &  0.848         & {\bf 0.805} & 0.869  &     0.822    & 0.870    & 0.867  \\
            \data{TMC2007}    &  {\bf 0.718}         & N/A    & 0.816  &  { 0.796}    & DNF      & DNF     \\
            \data{MediaMill}  &  N/A           & N/A    & {\bf 0.878}  &  0.913    & DNF      & DNF    \\
		\bottomrule
	\end{tabular}\\
	\vspace{2mm}
	\textsc{Hamming Loss}\\
	\begin{tabular}{lllllll}
		\toprule
			           & \texttt{PCC-$\epsilon$} & \texttt{PCC-Beam}   & \texttt{RF-PCT} & \texttt{MCC} & \texttt{M$_\s$CC} & \texttt{P$_t$M$_\s$CC} \\
		\midrule                       
            \data{Music}       & 0.219          & 0.221 & \textbf{0.189}  &       0.226    & 0.216    & 0.202  \\ 
            \data{Scene}       & 0.107          & 0.106 & \textbf{0.094}  &    0.119    & 0.110     & 0.105  \\ 
            \data{Yeast}       & 0.211          & 0.210 & \textbf{0.197}  &    0.211    & 0.213    & 0.209  \\ 
            \data{Genbase}       & N/A            & \textbf{0.001} & N/A    &    {\bf 0.001}    & {\bf 0.001}    & {\bf 0.001}  \\ 
            \data{Medical}    & 0.015          & {\bf 0.011} & 0.014  & {\bf 0.011}       & {\bf 0.011}    & {\bf 0.011} \\
            \data{Enron}        & 0.046          & 0.052 & 0.046  &  0.053       & 0.58     & 0.58   \\
            \data{TMC2007}    & 0.055          & N/A    & {\bf 0.011}  & 0.076       & DNF      & DNF     \\
            \data{MediaMill}  & N/A            & N/A    & {\bf 0.029}  & 0.034       & DNF      & DNF    \\
		\bottomrule
	\end{tabular}\\
\end{table*}

\begin{table*}
{
	\footnotesize
	\centering 
	\caption{\label{table:accuracy}Predictive Performance from 5-fold CV, displayed as: value rank, i.e., the average value across all folds and the rank of that value for each dataset. Note that the rank is based on a higher precision than shown in the table.}
	\par{\eval{exact match}}\\
	\begin{tabular}{lrrrrrrrrr}
\toprule
Dataset   & \alg{IC}&\alg{CC}&\alg{PCC}&\alg{ECC}&\alg{CDN}&\alg{MCC}&\alg{M$_\s$CC}&\alg{PM$_\s$CC}&\alg{P$_t$M$_\s$CC} \\
\midrule
SolFlare  & 0.77 7 & 0.80 1 & 0.78 5 & 0.69 8 & 0.59 9 & 0.78 5 & 0.79 2 & 0.79 2 & 0.79 2  \\ 
Bridges & 0.09 9 & 0.12 4 & 0.12 4 & 0.10 8 & 0.14 1 & 0.12 4 & 0.13 2 & 0.13 2 & 0.11 7  \\ 
Parkins & 0.17 1 & 0.17 2 & 0.16 5 & 0.16 8 & 0.16 5 & 0.16 5 & 0.16 8 & 0.17 4 & 0.17 2  \\ 
Thyroid & 0.83 6 & 0.02 9 & 0.84 3 & 0.82 7 & 0.78 8 & 0.84 3 & 0.84 3 & 0.84 2 & 0.85 1  \\ 
Music     & 0.30 7 & 0.29 9 & 0.35 4 & 0.31 6 & 0.30 8 & 0.35 4 & 0.36 3 & 0.37 1 & 0.37 2  \\ 
Scene    & 0.54 8 & 0.55 7 & 0.64 3 & 0.61 6 & 0.53 9 & 0.64 3 & 0.63 5 & 0.68 2 & 0.69 1  \\ 
Yeast    & 0.14 7 & 0.15 6 & DNF     & 0.19 5 & 0.07 8 & 0.21 4 & 0.22 3 & 0.23 1 & 0.22 2  \\ 
Genbase  & 0.94 8 & 0.96 2 & DNF     & 0.94 6 & 0.94 6 & 0.96 2 & 0.96 2 & 0.96 5 & 0.96 1  \\ 
Medical   & 0.58 8 & 0.62 4 & DNF     & 0.64 1 & 0.60 7 & 0.63 2 & 0.62 3 & 0.62 5 & 0.60 6  \\ 
Enron  & 0.07 8 & 0.10 3 & DNF     & 0.11 1 & 0.07 7 & 0.10 2 & 0.09 6 & 0.09 5 & 0.10 4  \\ 
Reuters & 0.29 7 & 0.35 6 & DNF     & 0.36 5 & 0.27 8 & 0.37 4 & 0.37 1 & 0.37 1 & 0.37 3  \\ 
\midrule
avg. rank  & 6.91  & 4.82  & 4.00  & 5.55  & 6.91  & 3.45  & 3.45  & 2.73  & 2.82  \\
\bottomrule
\end{tabular}     
{\scriptsize \par Nemenyi signif.: \alg{MCC}$\succ$\alg{IC}; \alg{MCC}$\succ$\alg{CDN}; \alg{M$_\s$CC}$\succ$\alg{IC}; \alg{M$_\s$CC}$\succ$\alg{CDN}; \alg{PM$_\s$CC}$\succ$\alg{IC}; \alg{PM$_\s$CC}$\succ$\alg{CDN}; \alg{P$_t$M$_\s$CC}$\succ$\alg{IC}; \alg{P$_t$M$_\s$CC}$\succ$\alg{CDN}}
	
	\vspace{2mm}
	\par{\eval{Hamming score}}\\
	\begin{tabular}{lrrrrrrrrr}
\toprule
Dataset   & \alg{IC}&\alg{CC}&\alg{PCC}&\alg{ECC}&\alg{CDN}&\alg{MCC}&\alg{M$_\s$CC}&\alg{PM$_\s$CC}&\alg{P$_t$M$_\s$CC} \\
\midrule
SolFlare  & 0.90 7 & 0.92 1 & 0.90 4 & 0.85 8 & 0.77 9 & 0.90 4 & 0.90 6 & 0.90 2 & 0.90 2  \\ 
Bridges & 0.63 7 & 0.66 3 & 0.67 1 & 0.64 6 & 0.62 9 & 0.67 1 & 0.65 4 & 0.65 4 & 0.63 8  \\ 
Parkins & 0.68 2 & 0.68 1 & 0.67 7 & 0.68 3 & 0.68 3 & 0.67 7 & 0.67 7 & 0.68 3 & 0.68 6  \\ 
Thyroid & 0.97 1 & 0.83 9 & 0.97 1 & 0.97 7 & 0.96 8 & 0.97 1 & 0.97 6 & 0.97 1 & 0.97 1  \\ 
Music     & 0.81 4 & 0.79 8 & 0.80 6 & 0.81 5 & 0.79 9 & 0.80 6 & 0.81 3 & 0.81 1 & 0.81 1  \\ 
Scene    & 0.89 7 & 0.86 8 & 0.89 4 & 0.90 3 & 0.86 9 & 0.89 4 & 0.89 6 & 0.90 2 & 0.91 1  \\ 
Yeast    & 0.79 1 & 0.75 7 & DNF     & 0.79 3 & 0.72 8 & 0.78 6 & 0.79 4 & 0.79 2 & 0.79 4  \\ 
Genbase  & 1.00 7 & 1.00 1 & DNF     & 1.00 7 & 1.00 5 & 1.00 1 & 1.00 1 & 1.00 5 & 1.00 1  \\ 
Medical   & 0.99 5 & 0.99 2 & DNF     & 0.99 1 & 0.99 8 & 0.99 2 & 0.99 2 & 0.99 5 & 0.99 5  \\ 
Enron  & 0.93 2 & 0.92 5 & DNF     & 0.94 1 & 0.92 3 & 0.92 5 & 0.92 8 & 0.92 5 & 0.92 3  \\ 
Reuters & 0.98 1 & 0.98 1 & DNF     & 0.98 1 & 0.98 8 & 0.98 1 & 0.98 1 & 0.98 1 & 0.98 1  \\ 
\midrule
avg. rank  & 4.00  & 4.18  & 3.83  & 4.09  & 7.18  & 3.45  & 4.36  & 2.82  & 3.00  \\
\bottomrule
\end{tabular}     
{\scriptsize \par Nemenyi signif.: \alg{PCC}$\succ$\alg{CDN}; \alg{MCC}$\succ$\alg{CDN}; \alg{PM$_\s$CC}$\succ$\alg{CDN}; \alg{P$_t$M$_\s$CC}$\succ$\alg{CDN}}

}
\end{table*}

\begin{table}
	\centering 
{\scriptsize
	\caption{\label{table:rtime}Running time; averaged over 5-fold CV and rounded nearest second. 
		Dataset-wise rankings, and some of the smaller datasets are not shown due to space limitations.
	}

	\par{\eval{running time} (training + testing)}\\
	\begin{tabular}{lrrrrrrrrr}
\toprule
Dataset   & \alg{IC}&\alg{CC}&\alg{PCC}&\alg{ECC}&\alg{CDN}&\alg{MCC}&\alg{M$_\s$CC}&\alg{PM$_\s$CC}&\alg{P$_t$M$_\s$CC} \\
\midrule
Scene    & 12  & 11  & 15  & 44  & 92  & 90  & 1347  & 684  & 335   \\ 
Yeast    & 11  & 11  & DNF     & 66  & 88  & 149  & 1313  & 731  & 546   \\ 
Genbase  & 11  & 8  & DNF     & 56  & 573  & 1695  & 5287  & 774  & 823   \\ 
Medical   & 9  & 11  & DNF     & 86  & 1546  & 3420  & 6940  & 1038  & 1192   \\ 
Enron  & 102  & 92  & DNF     & 349  & 3091  & 3884  & 10821  & 2986  & 3470   \\ 
Reuters & 106  & 120  & DNF     & 20593  & 14735  & 1837  & 5740  & 4890  & 5310   \\ 
\toprule
\end{tabular}     

	\vspace{2mm}
	\par{\eval{build time} (training only)}\\
	\begin{tabular}{lrrrrrrrrr}
\toprule
Dataset   & \alg{IC}&\alg{CC}&\alg{PCC}&\alg{ECC}&\alg{CDN}&\alg{MCC}&\alg{M$_\s$CC}&\alg{PM$_\s$CC}&\alg{P$_t$M$_\s$CC} \\
\midrule
Scene    & 12  & 10  & 11  & 43  & 13  & 13  & 1233  & 671  & 322   \\ 
Yeast    & 11  & 11  & DNF     & 64  & 12  & 15  & 1164  & 707  & 525   \\ 
Genbase  & 9  & 7  & DNF     & 46  & 5  & 11  & 3875  & 651  & 683   \\ 
Medical   & 8  & 8  & DNF     & 63  & 7  & 12  & 4986  & 835  & 961   \\ 
Enron  & 99  & 86  & DNF     & 307  & 72  & 80  & 8474  & 2729  & 3139   \\ 
Reuters & 96  & 102  & DNF     & 2030  & 120  & 108  & 4920  & 4120  & 4449   \\ 
\bottomrule
\end{tabular}     

}
\end{table}

{\pink
As in the literature, \texttt{CC} improves over \texttt{IC} considerably, particularly under \eval{exact match}, where label dependence must be modelled for best performance.

\texttt{PCC} improves further on \texttt{CC} -- in the cases where it is tractable -- also across both evaluation measures.

In \Tab{table:accuracy} we see that \texttt{MCC} outperforms \texttt{CC} on almost every occasion (only two exceptions) and is identical to \texttt{PCC} on all datasets where \texttt{PCC} completes, indicating that our methods conduct accurate inference like \texttt{PCC}, but are much more computationally tractable. Recall that \texttt{PCC}'s inference is \emph{optimal}.

Overall, \texttt{M$_\s$CC} obtains similar performance to \texttt{MCC}. It would follow that higher performance could be obtained with a higher value of $T_s$, hinted at by the fact that most wins over \texttt{M$_\s$CC} are on the smaller datasets (\textsf{SolFlare, Bridges, Music}) where the chain-sequence space is smaller and easier to explore (with small $T_s$). Of course, increasing $T_s$ implies a correspondingly increased computational cost. On the other hand, it is likely that this issue stems from the fact that a single chain sequence may not necessarily be best for predicting \emph{all} test instances. This was a motivation behind our population-of-$\s$ method, \texttt{PM$_\s$CC}.


In \Tab{table:accuracy} \texttt{PM$_\s$CC} obtains the best performance of all methods, under both \eval{exact match} and \eval{Hamming loss}, across the majority of datasets. It appears that on \textsf{Medical}, \textsf{Enron} the random chains of \texttt{ECC} provide better performance. Perhaps adding an \texttt{ECC}-like voting scheme would make up the difference. 

\texttt{P$_t$M$_\s$CC} obtains almost as good performance as \texttt{PM$_\s$CC}, but is more efficient on most datasets. This is exactly what it was designed it be: an efficient version of \texttt{PM$_\s$CC}. However, we were surprised to see in \Tab{table:rtime} that it is not more efficient on some of the large datasets, indicating that there may be some overhead in our implementation which is sensitive to $L$.
}

In \Tab{table:new} competition to \texttt{P$_t$M$_\s$CC} is shown by \texttt{PCC-Beam} under \eval{exact match}, and \texttt{RF-PCT} under \eval{Hamming loss}. It is {\pink worth noting that the Beam-search implementation paid more attention to setting up the base classifier, and we select the $_5^1$ configuration from \cite{BeamSearch} as the best of several combinations, whereas we used a single combination of $T_y$/$T_s$/$M$ parameters in our MCC methods. Both chaining approaches are quite competitive, especially taking into account that \texttt{RF-PCT} was pegged one of the best of 12 methods in \cite{ExtML}'s evaluation. It is true that \texttt{PF-PCT} is more efficient. The \texttt{PCC-Beam} paper does not report results for larger datasets like \textsf{TMC2007} and \textsf{MediaMill}.} 

{\pink Although in \Tab{table:new} the methods which search the label sequence space do not finish within the 24 hour cut off, we point out that for \texttt{P$_t$M$_\s$CC} this is a less-than linear increase with $T_s$ (depending on $\beta$). With $T_s = 10$, we expect the method to take $<10$ times longer than \texttt{MCC}. Further speedups are possible, for example using faster classifiers and/or smaller sets for internal validation. We intend to investigate this in future work. Nevertheless, it is clear that searching the chain space in this fashion becomes less feasible for larger amounts of data.}

Our methods are generally faster than \texttt{CDN}, especially for larger $L$. This is interesting, since an attraction of `\textit{chain-less}' methods like \texttt{CDN} is that no study of chain sequence is necessary.  However, we see that in this case, although there is no need to choose a chain-sequence, inference is relatively much more costly. 

As a side note, we point out that there is clearly a qualitative difference between the multi-dimensional datasets where $K_\l > 2$, and the binary-labelled datasets, where $K_\l = 2$. For example on \textsf{Bridges} the otherwise-more-advanced methods (\texttt{ECC} and \texttt{MCC} and variations thereof) perform relatively poorly compared to the basic baseline \texttt{IC}, which performs best. {\pink More than anything this is probably due to the relatively smaller size of these datasets, making it difficult to get a good approximation $\hat{p}(\y|\x) \approx p(\y|\x)$ of the true {\red{density}}. In future work we intend to create larger MDC ($K_\l > 2$) datasets and more thoroughly investigate this.}

%

\section{Conclusions and future work}
\label{sec:Concl}

We designed novel Monte Carlo (MC) schemes to perform efficient inference in a multi-dimensional learning framework using classifier chains.
MC techniques are used to efficiently search the chain-order space at the training stage, and the label-path space at the inference stage.
%
%
We analysed in detail the different possible choices of payoff functions for these MC methods, both from a practical and theoretical point of view.

An extensive empirical evaluation showed that using our techniques yields better predictive performance than many related methods while remaining computationally tractable.
Our model convincingly obtains overall best predictive performance of all the methods we looked at, and proves tractable enough for many real-world applications.

In future work, we intend to look at more advanced search algorithms and dependency structures other than chain models.

\section*{Acknowledgements}
\label{sec:acknowledgements}
{\footnotesize
This work has been partly supported by the Spanish government through projects COMONSENS (CSD2008-00010), ALCIT (TEC2012-38800-C03-01), DISSECT (TEC2012-38058-C03-01) and COMPREHENSION (TEC2012-38883-C02-01).
}

\appendix

\section{Additive Cost function for the Hamming Loss}
\label{sec:costHamming}

Inserting \eqref{eq:hammingLoss} into \eqref{eq:labelOrderAdditive2}, we notice that the optimum label order for the additive cost function with the Hamming loss is given by
{\footnotesize
\begin{align}
	\hat{\s}_{\textrm{Ham}}(\mat{X}) = & \argmin_{\s \in \S^L} \sum_{n=1}^{N}{\sum_{\y_{\s}^{(n)} \in {\boldsymbol \Y}}
			{\sum_{\ell=1}^{L}{\llbracket y_{s_{\l}}^{(n)} \ne \hat{y}_{s_{\l}}^{(n)} \rrbracket \hat{p}(\y_{\s}^{(n)}|\x^{(n)})}}} \nonumber \\
		= & \argmin_{\s \in \S^L}
			\sum_{n=1}^{N} \sum_{\ell=1}^{L} \sum_{y_{s_{\l}}^{(n)} \in \Y_{s_{\l}}}{
				\llbracket y_{s_{\l}}^{(n)} \ne \hat{y}_{s_{\l}}^{(n)} \rrbracket {\pink{\hat{p}(y_{s_{\l}}^{(n)}|\x^{(n)})}}} \nonumber \\
		 & \times \sum_{\y_{\s \neg s_{\l}}^{(n)} \in {\boldsymbol \Y} \setminus \Y_{s_{\l}}}{
		 	\prod_{\substack{k=1 \\ k \ne \ell}}^{{\red{L}}}{\hat{p}(y_{s_{k}}^{(n)}|\x^{(n)},{\red{y_{s_{\l}}^{(n)}}},y_{s_{1}}^{(n)},\ldots,y_{s_{k-1}}^{(n)})}},
\label{eq:labelOrderHamming2}
\end{align}
}
where {\red{we have used the chain rule of probability with $y_{s_{\l}}^{(n)}$ as root node, $\y_{\s \neg s_{\l}}^{(n)} = [y_{s_{1}}^{(n)},\ldots,y_{s_{\l-1}}^{(n)},y_{s_{\l+1}}^{(n)},\ldots,y_{s_{L}}^{(n)}]^{\top}$, and the last expression has been obtained simply separating the $\l$-th label from the rest.}}
Now, noticing that the last term in \eqref{eq:labelOrderHamming2} is equal to one, we obtain Eq. \eqref{eq:labelOrderHamming}:
{\footnotesize
\begin{align}
	\hat{\s}_{\textrm{Ham}}(\mat{X}) & = \argmin_{\s \in \S^L}
			\sum_{n=1}^{N} \sum_{\l=1}^{L} \sum_{y_{s_{\l}}^{(n)} \in \Y_{s_{\l}}}{
				\llbracket y_{s_{\l}}^{(n)} \ne \hat{y}_{s_{\l}}^{(n)} \rrbracket {\pink{\hat{p}(y_{s_{\l}}^{(n)}|\x^{(n)})}}} \nonumber \\
		& = \argmin_{\s \in \S^L} \sum_{n=1}^{N}{\sum_{\l=1}^{L}{\left(1-{\pink{\hat{p}(\hat{y}_{s_{\l}}^{(n)}|\x^{(n)})}}\right)}}
			\nonumber \\
		& = \argmax_{\s \in \S^L} \sum_{n=1}^{N}{\sum_{\l=1}^{L}{{\pink{\hat{p}(\hat{y}_{s_{\l}}^{(n)}|\x^{(n)})}}}}.
\label{eq:labelOrderHamming3}
\end{align}
}



\bibliographystyle{model1-num-names}
\bibliography{MCMC,SSP}

\begin{thebibliography}{29}
\expandafter\ifx\csname natexlab\endcsname\relax\def\natexlab#1{#1}\fi
\providecommand{\bibinfo}[2]{#2}
\ifx\xfnm\relax \def\xfnm[#1]{\unskip,\space#1}\fi
\bibitem[{Bielza et~al.(2011)Bielza, Li, and Larra{\~n}aga}]{Bielza2011}
\bibinfo{author}{C.~Bielza}, \bibinfo{author}{G.~Li},
  \bibinfo{author}{P.~Larra{\~n}aga},
\newblock \bibinfo{title}{Multi-dimensional classification with {B}ayesian
  networks},
\newblock \bibinfo{journal}{I. Journal of Approximate Reasoning}
  \bibinfo{volume}{52} (\bibinfo{year}{2011}) \bibinfo{pages}{705--727}.
\bibitem[{Kocev et~al.(2013)Kocev, Vens, Struyf, and Deroski}]{RFPCT}
\bibinfo{author}{D.~Kocev}, \bibinfo{author}{C.~Vens},
  \bibinfo{author}{J.~Struyf}, \bibinfo{author}{S.~Deroski},
\newblock \bibinfo{title}{Tree ensembles for predicting structured outputs},
\newblock \bibinfo{journal}{Pattern Recognition} \bibinfo{volume}{46}
  (\bibinfo{year}{2013}) \bibinfo{pages}{817--833}.
\bibitem[{Kocev et~al.(2007)Kocev, Vens, Struyf, and D\v{z}eroski}]{MODTs}
\bibinfo{author}{D.~Kocev}, \bibinfo{author}{C.~Vens},
  \bibinfo{author}{J.~Struyf}, \bibinfo{author}{S.~D\v{z}eroski},
\newblock \bibinfo{title}{Ensembles of multi-objective decision trees},
\newblock in: \bibinfo{booktitle}{Proceedings of the 18th European conference
  on Machine Learning}, ECML '07, \bibinfo{publisher}{Springer-Verlag},
  \bibinfo{address}{Berlin, Heidelberg}, \bibinfo{year}{2007}, pp.
  \bibinfo{pages}{624--631}.
\bibitem[{Tsoumakas and Katakis(2007)}]{Overview}
\bibinfo{author}{G.~Tsoumakas}, \bibinfo{author}{I.~Katakis},
\newblock \bibinfo{title}{Multi label classification: An overview},
\newblock \bibinfo{journal}{International Journal of Data Warehousing and
  Mining} \bibinfo{volume}{3} (\bibinfo{year}{2007}) \bibinfo{pages}{1--13}.
\bibitem[{Carvalho and Freitas(2009)}]{Carvalho2009}
\bibinfo{author}{A.~C. Carvalho}, \bibinfo{author}{A.~A. Freitas},
\newblock \bibinfo{title}{A tutorial on multi-label classification techniques},
\newblock in: \bibinfo{editor}{A.~Abraham}, \bibinfo{editor}{A.-E. Hassanien},
  \bibinfo{editor}{V.~Snášel} (Eds.), \bibinfo{booktitle}{Foundations of
  Computational Intelligence Volume 5}, volume \bibinfo{volume}{205} of
  \textit{\bibinfo{series}{Studies in Computational Intelligence}},
  \bibinfo{publisher}{Springer}, \bibinfo{year}{2009}, pp.
  \bibinfo{pages}{177--195}.
\bibitem[{Read(2010)}]{Thesis}
\bibinfo{author}{J.~Read}, \bibinfo{title}{Scalable Multi-label
  Classification}, Ph.D. thesis, University of Waikato, \bibinfo{year}{2010}.
\bibitem[{Tsoumakas et~al.(2010)Tsoumakas, Katakis, and Vlahavas}]{MMD}
\bibinfo{author}{G.~Tsoumakas}, \bibinfo{author}{I.~Katakis},
  \bibinfo{author}{I.~Vlahavas},
\newblock \bibinfo{title}{Mining multi-label data},
\newblock in: \bibinfo{editor}{O.~Maimon}, \bibinfo{editor}{L.~Rokach} (Eds.),
  \bibinfo{booktitle}{Data Mining and Knowledge Discovery Handbook},
  \bibinfo{publisher}{2nd edition, Springer}, \bibinfo{year}{2010}.
\bibitem[{Boutell et~al.(2004)Boutell, Luo, Shen, and Brown}]{Scene}
\bibinfo{author}{M.~R. Boutell}, \bibinfo{author}{J.~Luo},
  \bibinfo{author}{X.~Shen}, \bibinfo{author}{C.~M. Brown},
\newblock \bibinfo{title}{Learning multi-label scene classification},
\newblock \bibinfo{journal}{Pattern Recognition} \bibinfo{volume}{37}
  (\bibinfo{year}{2004}) \bibinfo{pages}{1757--1771}.
\bibitem[{Qi et~al.(2009)Qi, Hua, Rui, Tang, and Zhang}]{Qi2009}
\bibinfo{author}{G.-J. Qi}, \bibinfo{author}{X.-S. Hua},
  \bibinfo{author}{Y.~Rui}, \bibinfo{author}{J.~Tang}, \bibinfo{author}{H.-J.
  Zhang},
\newblock \bibinfo{title}{Two-dimensional multilabel active learning with an
  efficient online adaptation model for image classification},
\newblock \bibinfo{journal}{IEEE Trans. on Pattern Analysis and Machine
  Intelligence} \bibinfo{volume}{31} (\bibinfo{year}{2009})
  \bibinfo{pages}{1880--1897}.
\bibitem[{Zhang and Zhou(2006)}]{BPMLL}
\bibinfo{author}{M.-L. Zhang}, \bibinfo{author}{Z.-H. Zhou},
\newblock \bibinfo{title}{Multilabel neural networks with applications to
  functional genomics and text categorization},
\newblock \bibinfo{journal}{IEEE Transactions on Knowledge and Data
  Engineering} \bibinfo{volume}{18} (\bibinfo{year}{2006})
  \bibinfo{pages}{1338--1351}.
\bibitem[{Trohidis et~al.(2008)Trohidis, Tsoumakas, Kalliris, and
  Vlahavas}]{Music}
\bibinfo{author}{K.~Trohidis}, \bibinfo{author}{G.~Tsoumakas},
  \bibinfo{author}{G.~Kalliris}, \bibinfo{author}{I.~Vlahavas},
\newblock \bibinfo{title}{Multilabel classification of music into emotions},
\newblock in: \bibinfo{booktitle}{ISMIR '08: 9th International Conference on
  Music Information Retrieval}.
\bibitem[{Barutcuoglu et~al.(2006)Barutcuoglu, Schapire, and
  Troyanskaya}]{GeneFunction}
\bibinfo{author}{Z.~Barutcuoglu}, \bibinfo{author}{R.~E. Schapire},
  \bibinfo{author}{O.~G. Troyanskaya},
\newblock \bibinfo{title}{Hierarchical multi-label prediction of gene
  function},
\newblock \bibinfo{journal}{Bioinformatics} \bibinfo{volume}{22}
  (\bibinfo{year}{2006}) \bibinfo{pages}{830--836}.
\bibitem[{Hastie and Tibshirani(1998)}]{SMOM}
\bibinfo{author}{T.~Hastie}, \bibinfo{author}{R.~Tibshirani},
\newblock \bibinfo{title}{Classification by pairwise coupling},
\newblock in: \bibinfo{editor}{M.~I. Jordan}, \bibinfo{editor}{M.~J. Kearns},
  \bibinfo{editor}{S.~A. Solla} (Eds.), \bibinfo{booktitle}{Advances in Neural
  Information Processing Systems (NIPS)}, volume~\bibinfo{volume}{10},
  \bibinfo{publisher}{MIT Press}, \bibinfo{year}{1998}.
\bibitem[{Tsoumakas and Vlahavas(2007)}]{RAKEL}
\bibinfo{author}{G.~Tsoumakas}, \bibinfo{author}{I.~P. Vlahavas},
\newblock \bibinfo{title}{Random k-labelsets: An ensemble method for multilabel
  classification},
\newblock in: \bibinfo{booktitle}{ECML '07: 18th European Conference on Machine
  Learning}, \bibinfo{publisher}{Springer}, \bibinfo{year}{2007}, pp.
  \bibinfo{pages}{406--417}.
\bibitem[{Read et~al.(2011)Read, Pfahringer, Holmes, and Frank}]{ECC2}
\bibinfo{author}{J.~Read}, \bibinfo{author}{B.~Pfahringer},
  \bibinfo{author}{G.~Holmes}, \bibinfo{author}{E.~Frank},
\newblock \bibinfo{title}{Classifier chains for multi-label classification},
\newblock \bibinfo{journal}{Machine Learning} \bibinfo{volume}{85}
  (\bibinfo{year}{2011}) \bibinfo{pages}{333--359}.
\bibitem[{Cheng et~al.(2010)Cheng, Dembczy{\'n}ski, and H\"ullermeier}]{PCC}
\bibinfo{author}{W.~Cheng}, \bibinfo{author}{K.~Dembczy{\'n}ski},
  \bibinfo{author}{E.~H\"ullermeier},
\newblock \bibinfo{title}{Bayes optimal multilabel classification via
  probabilistic classifier chains},
\newblock in: \bibinfo{booktitle}{27th International Conference on Machine
  Learning (ICML)}, \bibinfo{address}{Haifa, Israel}.
\bibitem[{Guo and Gu(2011)}]{GuoGu}
\bibinfo{author}{Y.~Guo}, \bibinfo{author}{S.~Gu},
\newblock \bibinfo{title}{Multi-label classification using conditional
  dependency networks.},
\newblock in: \bibinfo{booktitle}{Proc. of the 24th Int. Joint Conf. on
  Artificial Intelligence (IJCAI)}, pp. \bibinfo{pages}{1300--1305}.
\bibitem[{Zaragoza et~al.(2011)Zaragoza, Sucar, Morales, Bielza, and
  Larra{\~n}aga}]{UPM}
\bibinfo{author}{J.~H. Zaragoza}, \bibinfo{author}{L.~E. Sucar},
  \bibinfo{author}{E.~F. Morales}, \bibinfo{author}{C.~Bielza},
  \bibinfo{author}{P.~Larra{\~n}aga},
\newblock \bibinfo{title}{Bayesian chain classifiers for multidimensional
  classification},
\newblock in: \bibinfo{booktitle}{Proc. of the 24th Int. Joint Conf. on
  Artificial Intelligence (IJCAI)}.
\bibitem[{Read et~al.(2008)Read, Pfahringer, and Holmes}]{EPS}
\bibinfo{author}{J.~Read}, \bibinfo{author}{B.~Pfahringer},
  \bibinfo{author}{G.~Holmes},
\newblock \bibinfo{title}{Multi-label classification using ensembles of pruned
  sets},
\newblock in: \bibinfo{booktitle}{ICDM'08: Eighth IEEE International Conference
  on Data Mining}, \bibinfo{publisher}{IEEE}, \bibinfo{year}{2008}, pp.
  \bibinfo{pages}{995--1000}.
\bibitem[{Madjarov et~al.(2012)Madjarov, Kocev, Gjorgjevikj, and
  Deroski}]{ExtML}
\bibinfo{author}{G.~Madjarov}, \bibinfo{author}{D.~Kocev},
  \bibinfo{author}{D.~Gjorgjevikj}, \bibinfo{author}{S.~Deroski},
\newblock \bibinfo{title}{An extensive experimental comparison of methods for
  multi-label learning},
\newblock \bibinfo{journal}{Pattern Recognition} \bibinfo{volume}{45}
  (\bibinfo{year}{2012}) \bibinfo{pages}{3084--3104}.
\bibitem[{Dembczy{\'n}ski et~al.(2012)Dembczy{\'n}ski, Waegeman, Cheng, and
  H\"ullermeier}]{OnLabelDependenceJournal}
\bibinfo{author}{K.~Dembczy{\'n}ski}, \bibinfo{author}{W.~Waegeman},
  \bibinfo{author}{W.~Cheng}, \bibinfo{author}{E.~H\"ullermeier},
\newblock \bibinfo{title}{On label dependence and loss minimization in
  multi-label classification},
\newblock \bibinfo{journal}{Machine Learning} \bibinfo{volume}{88}
  (\bibinfo{year}{2012}) \bibinfo{pages}{5--45}.
\bibitem[{Kumar et~al.(2012)Kumar, Vembu, Menon, and Elkan}]{BeamSearch}
\bibinfo{author}{A.~Kumar}, \bibinfo{author}{S.~Vembu}, \bibinfo{author}{A.~K.
  Menon}, \bibinfo{author}{C.~Elkan},
\newblock \bibinfo{title}{Learning and inference in probabilistic classifier
  chains with beam search},
\newblock in: \bibinfo{booktitle}{Machine Learning and Knowledge Discovery in
  Databases}, volume \bibinfo{volume}{7523}, \bibinfo{publisher}{Springer},
  \bibinfo{year}{2012}, pp. \bibinfo{pages}{665--680}.
\bibitem[{Dembczy{\'n}ski et~al.(2012)Dembczy{\'n}ski, Waegeman, and
  H\"ullermeier}]{ChainingMLC}
\bibinfo{author}{K.~Dembczy{\'n}ski}, \bibinfo{author}{W.~Waegeman},
  \bibinfo{author}{E.~H\"ullermeier},
\newblock \bibinfo{title}{An analysis of chaining in multi-label
  classification},
\newblock in: \bibinfo{booktitle}{Workshop Proceedings of 20th European
  Conference on Artificial Intelligence (ECAI)}, \bibinfo{address}{Montpellier,
  France}, pp. \bibinfo{pages}{294--299}.
\bibitem[{Read et~al.(2013)Read, Martino, and Luengo}]{MCC}
\bibinfo{author}{J.~Read}, \bibinfo{author}{L.~Martino},
  \bibinfo{author}{D.~Luengo},
\newblock \bibinfo{title}{Efficient {M}onte {C}arlo optimization for
  multi-label classifier chains},
\newblock in: \bibinfo{booktitle}{To appear in ICASSP '13: 38th International
  Conference on Acoustics, Speech, and Signal Processing.}
\bibitem[{Dembczynski et~al.(2011)Dembczynski, Waegeman, Cheng, and
  H\"ullermeier}]{FmeasureMLC}
\bibinfo{author}{K.~J. Dembczynski}, \bibinfo{author}{W.~Waegeman},
  \bibinfo{author}{W.~Cheng}, \bibinfo{author}{E.~H\"ullermeier},
\newblock \bibinfo{title}{An exact algorithm for {F}-measure maximization},
\newblock in: \bibinfo{editor}{J.~Shawe-Taylor}, \bibinfo{editor}{R.~Zemel},
  \bibinfo{editor}{P.~Bartlett}, \bibinfo{editor}{F.~Pereira},
  \bibinfo{editor}{K.~Weinberger} (Eds.), \bibinfo{booktitle}{Advances in
  Neural Information Processing Systems (NIPS) 24}, \bibinfo{year}{2011}, pp.
  \bibinfo{pages}{1404--1412}.
\bibitem[{Kirkpatrick et~al.(1983)Kirkpatrick, Gelatt, and Vecchi}]{SA}
\bibinfo{author}{S.~Kirkpatrick}, \bibinfo{author}{C.~Gelatt},
  \bibinfo{author}{M.~Vecchi},
\newblock \bibinfo{title}{Optimisation by simulated annealing},
\newblock \bibinfo{journal}{Science} \bibinfo{volume}{220}
  (\bibinfo{year}{1983}) \bibinfo{pages}{671--680}.
\bibitem[{Trees(2001)}]{VanTrees:detectionEstimationLinearModulation2001}
\bibinfo{author}{H.~L.~V. Trees}, \bibinfo{title}{Detection, Estimation, and
  Modulation Theory. Part I: Detection, Estimation, and Linear Modulation
  Theory}, \bibinfo{publisher}{John Wiley \& Sons}, \bibinfo{address}{New York,
  NY (USA)}, \bibinfo{year}{2001}.
\bibitem[{Hall et~al.(2009)Hall, Frank, Holmes, Pfahringer, Peter, and
  Witten}]{WEKA}
\bibinfo{author}{M.~Hall}, \bibinfo{author}{E.~Frank},
  \bibinfo{author}{G.~Holmes}, \bibinfo{author}{B.~Pfahringer},
  \bibinfo{author}{R.~Peter}, \bibinfo{author}{I.~H. Witten},
\newblock \bibinfo{title}{The {WEKA} data mining software: An update},
\newblock \bibinfo{journal}{SIGKDD Explorations} \bibinfo{volume}{11}
  (\bibinfo{year}{2009}).
\bibitem[{Dem\v{s}ar(2006)}]{Nemenyi}
\bibinfo{author}{J.~Dem\v{s}ar},
\newblock \bibinfo{title}{Statistical comparisons of classifiers over multiple
  data sets},
\newblock \bibinfo{journal}{Journal of Machine Learning Research}
  \bibinfo{volume}{7} (\bibinfo{year}{2006}) \bibinfo{pages}{1--30}.

\end{thebibliography}







\end{document}